\documentclass[lettersize,journal]{IEEEtran}
\usepackage{amsmath,amsfonts}
\usepackage{algorithmic}
\usepackage{algorithm}
\usepackage{array}
\usepackage[caption=false,font=normalsize,labelfont=sf,textfont=sf]{subfig}
\usepackage{textcomp}
\usepackage{stfloats}
\usepackage{url}
\usepackage{verbatim}
\usepackage{graphicx}
\usepackage{cite}
\usepackage{xspace}
\usepackage{amsmath}
\usepackage{mathtools}
\usepackage[utf8]{inputenc}
\usepackage{bm}
\usepackage{booktabs}
\usepackage{multirow}
\usepackage{float}
\usepackage{animate}
\usepackage[dvipsnames]{xcolor}
\DeclareUnicodeCharacter{FF1A}{:}
\newcommand{\bigO}[1]{\mathcal{O}#1}

\usepackage{xspace}
\makeatletter
\DeclareRobustCommand\onedot{\futurelet\@let@token\@onedot}
\def\@onedot{\ifx\@let@token.\else.\null\fi\xspace}
\def\eg{\emph{e.g}\onedot} 
 
\def\cf{\emph{c.f}\onedot} 
 
\def\wrt{w.r.t\onedot}

\newcommand{\sg}[1]{\mathrm{sg}#1}
\newcommand{\sigmoid}[1]{\mathrm{sigmoid}#1}
\usepackage{xcolor}

\begin{document}

\title{DaRePlane: Direction-aware Representations\\for Dynamic Scene Reconstruction}

\author{Ange Lou\textsuperscript{1,2\thanks{This work was partly carried out during the internship of Ange Lou, Tianyu Luan, and Hao Ding at United Imaging Intelligence, Boston MA, USA.}}, Benjamin Planche\textsuperscript{2}\thanks{Corresponding authors: Benjamin Planche and Jack Noble \\({\tt\scriptsize benjamin.planche@uii-ai.com}, {\tt\scriptsize jack.noble@vanderbilt.edu}).}, Zhongpai Gao\textsuperscript{2}, Yamin Li\textsuperscript{1}, Tianyu Luan\textsuperscript{2,3}, Hao Ding\textsuperscript{2,4}, Meng Zheng, Terrence Chen\textsuperscript{2}, Ziyan Wu\textsuperscript{2}, Jack Noble\textsuperscript{1}\\
\textsuperscript{1}Vanderbilt University, Nashville TN, USA\\
\textsuperscript{2}United Imaging Intelligence, Boston MA, USA\\
\textsuperscript{3}Johns Hopkins University, Baltimore MD, USA\\
\textsuperscript{4}University of Buffalo, Buffalo NY, USA\\
{\tt\small {first.last}@vanderbilt.edu, {first.last}@uii-ai.com, tianyulu@buffalo.edu, hding15@jhu.edu}\\
}

\markboth{Preprint}%
{Lou \MakeLowercase{\textit{et al.}}: A Sample Article Using IEEEtran.cls for IEEE Journals}




\maketitle
\begin{figure*}[!htbp]
    \centering
    \includegraphics[width=\textwidth]{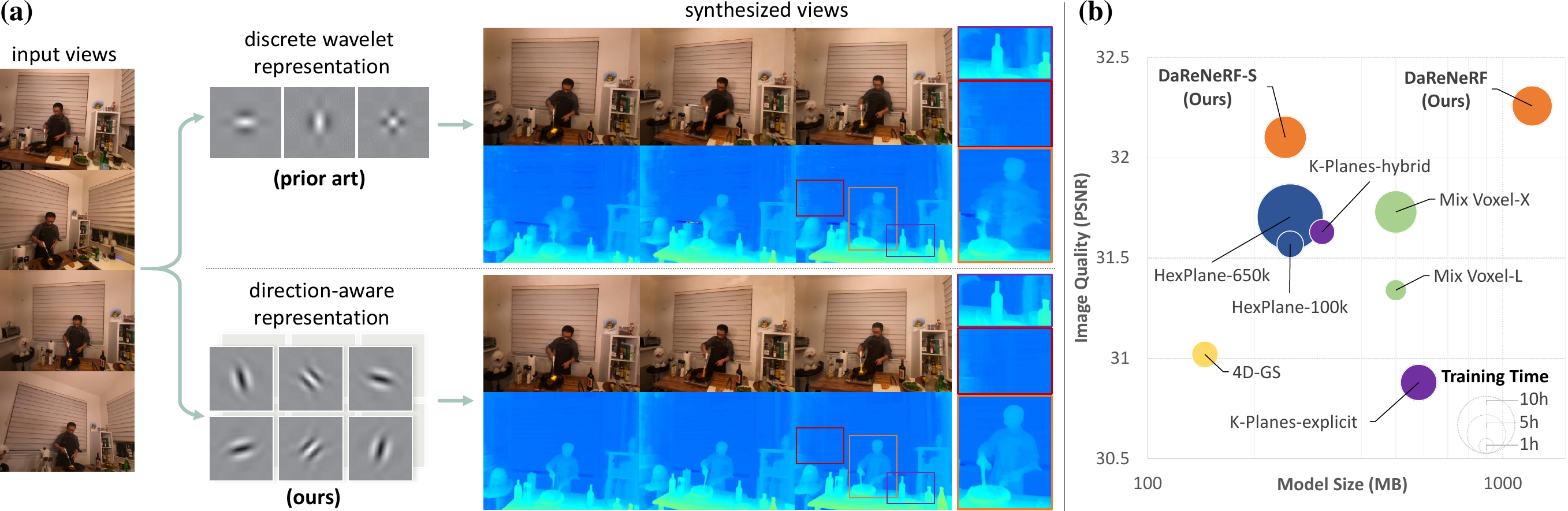} 
    \caption{\textbf{Performance of dynamic NeRF and Gaussian splatting (GS) with DaRePlane on 4D scenes.} Our direction-aware representation excels by capturing features of dynamic scenes from six different directions—a capability beyond the reach of traditional discrete-wavelet representations, \cf sub-figure (a). Built upon this advanced representation, 
    our NeRF method first introduced in \cite{lou2024darenerf} outperforms prior work in challenging 4D scenarios while being competitive in terms of training time and model size, offering the best trade-off overall, \cf sub-figure (b). Similar results for our GS solution are shared in Figure \ref{fig:GS_dareplane}.
    }
    \label{fig:figure_1}
\end{figure*}

\begin{abstract}
Numerous recent approaches to modeling and re-rendering dynamic scenes leverage plane-based explicit representations, addressing slow training times associated with models like neural radiance fields (NeRF) and Gaussian splatting (GS). 
However, merely decomposing 4D dynamic scenes into multiple 2D plane-based representations is insufficient for high-fidelity re-rendering of scenes with complex motions. In response, we present DaRePlane, a novel direction-aware representation approach that captures scene dynamics from six different directions. This learned representation undergoes an inverse dual-tree complex wavelet transformation (DTCWT) to recover plane-based information. 
Within NeRF pipelines, DaRePlane computes features for each space-time point by fusing vectors from these recovered planes, then passed to a tiny MLP for color regression.
When applied to Gaussian splatting, DaRePlane computes the features of Gaussian points, followed by a tiny multi-head MLP for spatial-time deformation prediction. Notably, to address redundancy introduced by the six real and six imaginary direction-aware wavelet coefficients, we introduce a trainable masking approach, mitigating storage issues without significant performance decline. To demonstrate the generality and efficiency of DaRePlane, we test it on both regular and surgical dynamic scenes, for both NeRF and GS systems. Extensive experiments show that DaRePlane yields state-of-the-art performance in novel view synthesis for various complex dynamic scenes. 
\end{abstract}


\section{Introduction}
\IEEEPARstart{T}{he} reconstruction and re-rendering of 3D scenes from a set of 2D images pose a fundamental challenge in computer vision, holding substantial implications for a range of AR/VR applications \cite{wang2022neural,moreau2022lens,wysocki2023ultra}. 
Despite recent progress in reconstructing static scenes, significant challenges remain. 
Real-world scenes are inherently dynamic, characterized by intricate motion, further adding to the task complexity. 

Recent popular reconstruction approaches can be summarized into two main categories: neural radiance fields (NeRF) \cite{mildenhall2021nerf} and Gaussian splatting (GS) \cite{kerbl20233d}. NeRF-related methods are known for achieving high-fidelity reconstruction performance, capturing fine details and complex scene geometries. NeRF works by formulating a scene as a continuous volumetric field, where each point in space (static) or space-time (dynamic) has a corresponding color and density. This information is then rendered using a differentiable volumetric rendering process. However, these methods suffer from extensive optimization times and low inference speeds. In contrast, GS represents a scene using an explicit cloud of point-like Gaussians and employs a real-time differentiable renderer. This significantly reduces both optimization and novel-view synthesis times, making GS a more practical choice for real-time applications.

Recent dynamic scene reconstruction methods build on NeRF's implicit representation. Some utilize a large MLP to process spatial and temporal point positions, generating color outputs \cite{li2022neural,li2021neural,wang2021ibrnet}. Others aim to disentangle scene motion and appearance \cite{gao2021dynamic,park2021nerfies,park2021hypernerf,pumarola2021d,liu2023robust}. However, both approaches face computational challenges, requiring extensive MLP evaluations for novel view rendering. The slow training process, often spanning days or weeks, and the reliance on additional supervision like depth maps \cite{li2021neural,li2023dynibar,liu2023robust} limit their widespread adoption for dynamic scene modeling. Several recent studies \cite{cao2023hexplane,fridovich2023k,shao2023tensor4d} have proposed decomposition-based methods to address the training time challenge. Similar techniques are also introduced in GS to model the temporal deformations of Gaussians for dynamic scenes \cite{wu20244d,Xu_2024_CVPR, ren2023dreamgaussian4d}. However, relying solely on decomposition limits both NeRF's and GS's ability to capture high-fidelity texture details.

Recent studies have explored the possibility of incorporating frequency information into NeRF \cite{rho2023masked, xu2023wavenerf, wang2022fourier, wu2023neural, yang2023freenerf} and GS \cite{Zhang_2024_CVPR}. These frequency-based representations demonstrate promising performance in static-scene rendering, particularly in recovering detailed information. However, there is limited exploration \wrt the ability of these methods to scale from static to dynamic scenes. Additionally, HexPlane \cite{cao2023hexplane} has noted a significant degradation in reconstruction performance when using wavelet coefficients as a basis. This limitation is inherent to wavelets themselves, and we delve into a detailed discussion in the following paragraph.

Traditional 2D discrete wavelet transform (DWT) employs low/high-pass real wavelets to decompose a 2D image or grid into approximation and detail wavelet coefficients across different scales. These coefficients offer an efficient representation of both global and local image details. However, there are two significant drawbacks hindering the successful application of 2D DWT-based representations to dynamic scenes. The first is the \textbf{shift variance} problem \cite{bradley2003shift}, where even a small shift in the input signal significantly disrupts the wavelets' oscillation pattern. In dynamic 3D scenes, shifts are more pronounced than in static scenarios due to factors such as multi-object motion, camera motion, reflections, and variations in illumination. Simple DWT wavelet representations struggle to handle such variability, yielding poor results in dynamic regions. Another critical issue is the \textbf{poor direction selectivity} \cite{kingsbury1999image} in DWT representations. A 2D DWT produces a checkered pattern that blends representations from $\pm45^\circ$, lacking directional selectivity, which is less effective for capturing lines and edges in images. Consequently, DWT-based representations fail to adequately model dynamic scenes, leading to results with noticeable ghosting artifacts around moving objects as shown in Figure \ref{fig:figure_1}.

In a preliminary publication \cite{lou2024darenerf}, we addressed
these key limitations of the discrete wavelet transform (DWT) by introducing an efficient and robust frequency-based representation designed to overcome the challenges of shift variance and lack of direction selectivity in modeling dynamic scenes. 
Inspired by the dual-tree complex wavelet transform (DTCWT) \cite{selesnick2005dual}, 
we proposed
a direction-aware representation, aiming to learn features from six distinct orientations without introducing the checkerboard pattern observed in DWT. Leveraging the properties of complex wavelet transforms, our approach ensures shift invariance within the representation. This direction-aware representation proves successful in modeling complex dynamic scenes, achieving state-of-the-art performance.

In the present article, we propose to generalize our DaRePlane representation and apply it
to both NeRF and GS systems, testing it on regular and surgical dynamic scenes, each presenting their own challenges. Additionally, to highlight the generalizability of our proposed method (aimed at 4D scenarios), we extend its application to modelling static 3D scenes. In this context, our proposed DaRePlane demonstrates high-fidelity reconstruction performance and efficient storage capabilities. This versatility underscores the efficacy of our approach not only in dynamic scenes but also in static environments, affirming its potential as a general representation utility across various scenarios.

In summary, our contributions are as follows:
\begin{itemize}
    \item We are the first to leverage DTCWT in NeRF optimization, introducing a direction-aware representation to address the shift-variance and direction-ambiguity shortcomings in DWT-based representations. DaReNeRF thereby outperforms prior decomposition-based methods in modeling complex dynamic scenes.
    
    \item We implement a trainable mask method for dynamic scene reconstruction, effectively resolving the storage limitations associated with the direction-aware representation. This adaptation ensures memory efficiency comparable with current state-of-the-art methods.
    \item We extend our direction-aware representation to static scene reconstruction, and experiments demonstrate that our proposed method outperforms other state-of-the-art approaches, achieving a superior trade-off between performance and model size.
    \item $[Extension-contribution]$ We further prove that our proposed representation can transfer to Gaussian splatting solutions (DaReGS), to similarly improve their modeling capability.
    \item $[Extension-contribution]$ To demonstrate the generalizability of our proposed method, we test DaReNeRF and DaReGS in various surgical scenarios, including microscopy, endoscopy, and laparoscopy. Experiments show that our method is effective for reconstruction tasks across different areas.
\end{itemize}

\section{Related Work}
\label{sec:related}


\subsection{Learnable Scene Representations}
\noindent \textbf{Neural Radiance Field.} 
Neural Radiance Fields (NeRF) represent three-dimensional scenes by approximating a radiance field using a neural network. This radiance field describes the color and density values for each sample point along a ray from a specific view direction. Novel views can be synthesized through the process of volume rendering \cite{mildenhall2021nerf}. NeRF \cite{mildenhall2021nerf} and its variants \cite{barron2021mip,barron2022mip,mildenhall2022nerf,niemeyer2022regnerf,low2023robust,wang2023f2,bian2023nope,yan2023nerf} show impressive results on novel view synthesis and many other application including 3D reconstruction \cite{martin2021nerf,zhang2022nerfusion,zhu2022nice,kobayashi2022decomposing}, semantic segmentation \cite{liu2023instance,mirzaei2023spin}, object detection \cite{hu2023nerf,xu2023nerf,xie2023pixel,xu2023mononerd},  generative model \cite{chan2022efficient,chan2021pi,xie2023high}, and 3D content creation \cite{metzer2023latent,wang2023learning,deng2023nerdi}.
However, implicit neural representations suffer from slow rendering
due to the numerous costly MLP evaluations required for each pixel. Various spatial-decomposition methods \cite{fridovich2022plenoxels,chen2022tensorf,attal2023hyperreel,chan2022efficient} have been proposed to address the challenge of training speed in static scenes. \\
\noindent \textbf{Gaussian Splatting.} As another answer to NeRF's costly optimization time and inference, 3D-GS has recently revolutionized the field of neural rendering. It employs a set of anisotropic 3D Gaussians, each parameterized by its position, covariance, color, and opacity, in order to explicitly represent a scene. To generate views, these 3D Gaussians are projected onto the camera's imaging plane and rendered using point-based volume rendering \cite{kerbl20233d}. Due to its compactness and rasterization speed, 3D-GS is applied to various scenarios, including 3D generation \cite{Chen_2024_CVPR,tang2023dreamgaussian,Liu_2024_CVPR}, autonomous driving \cite{zhou2024drivinggaussian, zhou2024hugs}, scene understanding \cite{qin2024langsplat}, and medical imaging \cite{cai2024radiative,nikolakakis2024gaspct,gao2024ddgs}.\\

\noindent \textbf{Extension to Dynamic Scenes.} Both NeRF and Gaussian Splatting (GS) can be extended to dynamic versions for modeling time-varying scenes. In the NeRF framework, one straightforward approach is to extend a static NeRF by introducing an additional time dimension \cite{pumarola2021d} or by incorporating a latent code \cite{guo2023forward,liu2023robust,li2023dynibar,wang2023flow}. While these methods demonstrate strong capabilities in modeling complex real-world dynamic scenes, they face a severely under-constrained problem that necessitates additional supervision—such as depth, optical flow, or dense observations—to achieve satisfactory results. The substantial model size and weeks-long training times associated with these approaches further hinder their real-world applicability. An alternative solution involves employing separate MLPs to represent the deformation field and a canonical field \cite{pumarola2021d,song2022pref,zhang2023deformtoon3d,yan2023nerf,johnson2023unbiased}. Here, the canonical field captures the static scene, while the deformation field learns coordinate mappings to the canonical space over time. Although this method offers improvements over the previous approach, it still demands significant training time.

In the GS setting, a more common method for depicting dynamic scenes involves using explicit plane-based representations to model the spatiotemporal deformation of 3D Gaussians \cite{wu20244d,duisterhof2023md,liu2024lgs,li2024st,lu20243d}. 

\subsection{Scene Decomposition}
\noindent \textbf{Plane-Based Representations.} Plane-based representations applied to dynamic scenes have first been proposed for NeRF methods \cite{cao2023hexplane, fridovich2023k,shao2023tensor4d}. These approaches aim to alleviate the lengthy training times associated with dynamic scenes while maintaining the ability to model their complexity. They decompose a 4D scene into plane-based representations and employ a compact MLP to aggregate features for volumetric rendering of resulting images. A similar plane-based representation has then been integrated into the 3D-GS system \cite{wu20234d}, to aggregate the spatial-temporal deformation features of 3D Gaussians. Subsequently, multiple tiny MLPs are employed to predict the time-variant deformation of both position and covariance. While plane-based representation significantly reduces training time and memory storage for dynamic NeRF, and enhances training time and inference speed for dynamic Gaussian Splatting, it still faces challenges in preserving detailed texture information during rendering.\\
\noindent \textbf{Wavelet Optimization.} To further enhance rendering quality, wavelet-based representations \cite{rho2023masked,xu2023wavenerf,saragadam2023wire} have gained significant attention for their ability to improve NeRF's capability in capturing fine texture details, due to their proficiency in recovering high-fidelity signals. However, there has been limited exploration of the potential of wavelet-based representations for dynamic scene modeling. Applying wavelet-based representations directly to plane-based methods can lead to a significant performance decay, as illustrated in Figure \ref{fig:figure_1}. Similar degradation is also reported by HexPlane \cite{cao2023hexplane}, highlighting the inherent limitations of wavelets, namely, shift variance and direction ambiguity. To overcome these limitations and build a more effective general wavelet-based representation for both NeRF and GS, we propose a direction-aware representation, which 
preserves the ability to detect detailed textures without requiring additional supervision, achieving state-of-the-art performance in real-world and surgical dynamic scene reconstruction.

\subsection{Application to Surgical Videos}
One of the most promising and impactful application areas for NeRF and GS is the reconstruction of dynamic surgical scenes from videos captured by surgical robots. Accurate reconstruction of surgical scenes from video is critical for precise image-guided surgery. Current NeRF-based methods \cite{wang2022neural,zha2023endosurf,lou2024samsnerf,yang2023neural,yang2024efficient} achieve superior reconstructions compared to traditional SLAM-based approaches. The use of plane-based representations in dynamic NeRF \cite{yang2023neural,yang2024efficient} significantly reduces training time to just a few minutes, thereby improving the feasibility of clinical applications. Moreover, plane-based 4D Gaussian Splatting (4D-GS) \cite{liu2024endogaussian} further minimizes training time and enables real-time inference. However, reconstructing surgical scenes poses the challenge of requiring high-fidelity anatomical reconstructions, which are crucial for accurate registration with pre-operative imaging or other intra-operative imaging modalities and for providing precise 3D feedback to the surgeon.

\begin{figure*}[ht]
    \centering
    \includegraphics[width=\textwidth]{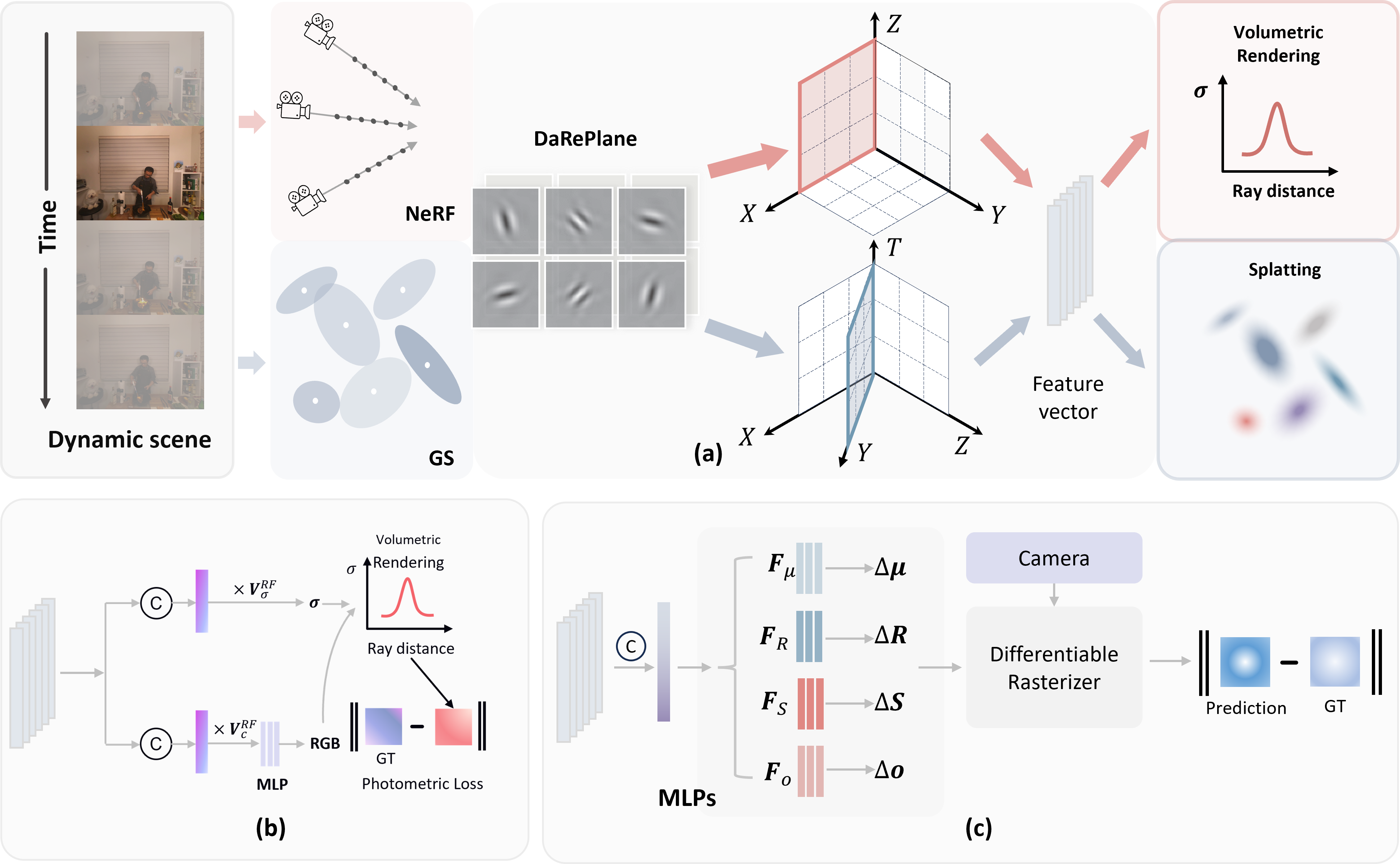} 
    \caption{\textbf{Method Overview.} \textbf{(a)} In the given sequence of images, NeRF and GS initialize the spatial-temporal points and a set of 3D Gaussians, respectively. Voxel features of these points (for NeRF) or Gaussians (for GS) are then computed by querying voxel planes in DaRePlane. These features are subsequently fed into the volumetric rendering process (for NeRF) or the splatting process (for GS) to synthesize the final images. \textbf{Bottom:} \textbf{(b) NeRF:} Feature vectors queried from DaRePlane are concatenated into a single vector, and then multiplies them by learned tensor $V^{RF}$ for final results. RGB colors are regressed by a compact MLP, and images are synthesized via volumetric rendering. \textbf{(c) GS:} The concatenated feature vector is decoded using a multi-head deformation decoder to obtain the deformation of Gaussians at a specific timestamp $t$. These deformed Gaussians are then splatted to render the final images.
    }
    \label{fig:dareplane_method}
\end{figure*}

\section{Method}
\label{sec:method}

We seek to develop a model for a dynamic scene using a collection of images captured from different viewpoints, each timestamped. The objective is to fit a model capable of rendering new images at varying poses and time stamps. Similar to D-NeRF \cite{pumarola2021d}, this model assigns color and opacity to points in both space and time. The rendering process involves differentiable volumetric rendering along rays. Training the entire model relies on a photometric loss function, comparing rendered images with ground-truth images to optimize model parameters.

Our primary innovation lies in introducing a novel direction-aware representation for dynamic scenes. This distinctive representation is coupled with the inverse dual-tree complex wavelet transform (IDTCWT) and a compact implicit multi-layer perceptron (MLP) to enable the generation of high-fidelity novel views. Figure \ref{fig:dareplane_method} 
shows an overview of the model. 
Note that for simplicity, we refer to the wavelet representation as wavelet coefficients in this section. 

\subsection{Plane-Based Representation}\label{sec:plane-rep}
A natural dynamic scene can be represented as a 4D spatio-temporal volume denoted as $D$. This 4D volume comprises individual static 3D volume for each time step, namely $\{V_1, V_2,...,V_T\}$. Directly modeling a 4D volume would entail a memory complexity of $\bigO(N^3TF)$, where $N$, $T$, $F$ are spatial resolution, temporal resolution and feature size (\eg, with $F=3$ representing RGB colors). To improve the overall performance, we propose a direction-aware representation applied to baseline plane-based 4D volume decomposition \cite{cao2023hexplane}. In such baseline, a representation of the 4D volume can be represented as follows:
\begin{small}
\begin{equation}
\begin{split}
    D=&\sum_{r=1}^{R_1}{M_r^{XY}\circ M_r^{ZT}\circ v_r^1}+\sum_{r=1}^{R_2}{M_r^{XZ}\circ M_r^{YT}\circ v_r^2} \\
    &+\sum_{r=1}^{R_3}{M_r^{YZ}\circ M_r^{XT}\circ v_r^3}
\end{split}
\end{equation}
\end{small}
where each $M_r^{AB}\in\mathbb{R}^{AB}$ represents a learned 2D plane-based representation with $\big\{(A,B) \in \{X, Y, Z, T\}^2 \mid A \ne B\big\}$, and $v_r^{i}\in\mathbb{R}^{F}$ are learned vectors along $F$ axes. The parameters $R_1$, $R_2$ and $R_3$ correspond to the number of low rank components. By defining  $R=R_1+R_2+R_3 \ll N$, the model's memory complexity can be notably reduced from $\bigO(N^3TF)$ to $\bigO(RN^2TF)$. This reduction in memory requirements proves advantageous for efficiently modeling dynamic scenes while preserving computational resources. 

Plane-based 4D NeRF models predict the density and appearance features of points in space-time by multiplying the feature vectors extracted from paired planes (\eg, $XY$ and $ZT$), concatenating the results into a single vector, and then multiplying them by $V^{RF}$.
The point opacities are directly queried from the density features, whereas the color values are regressed by a compact MLP conditioned on the appearance features and view directions. Finally, images are synthesized via volumetric rendering as shown in the \textbf{NeRF} setting of Figure \ref{fig:dareplane_method}. 

In \textbf{4D-GS} settings, the multiple feature vectors from the paired planes are also concatenated into a single feature vector, which is then processed through a multi-head MLP to predict the deformation of the Gaussians. Finally, the images are synthesized from the deformed Gaussians using a differentiable rasterizer, as shown in the \textbf{GS} setting of Figure \ref{fig:dareplane_method}. 

To improve the overall performance, we apply our proposed direction-aware representation to both NeRF and GS baselines.

\subsection{Direction-Aware Representation}
\begin{figure}
    \centering
    \includegraphics[width=1.\linewidth]{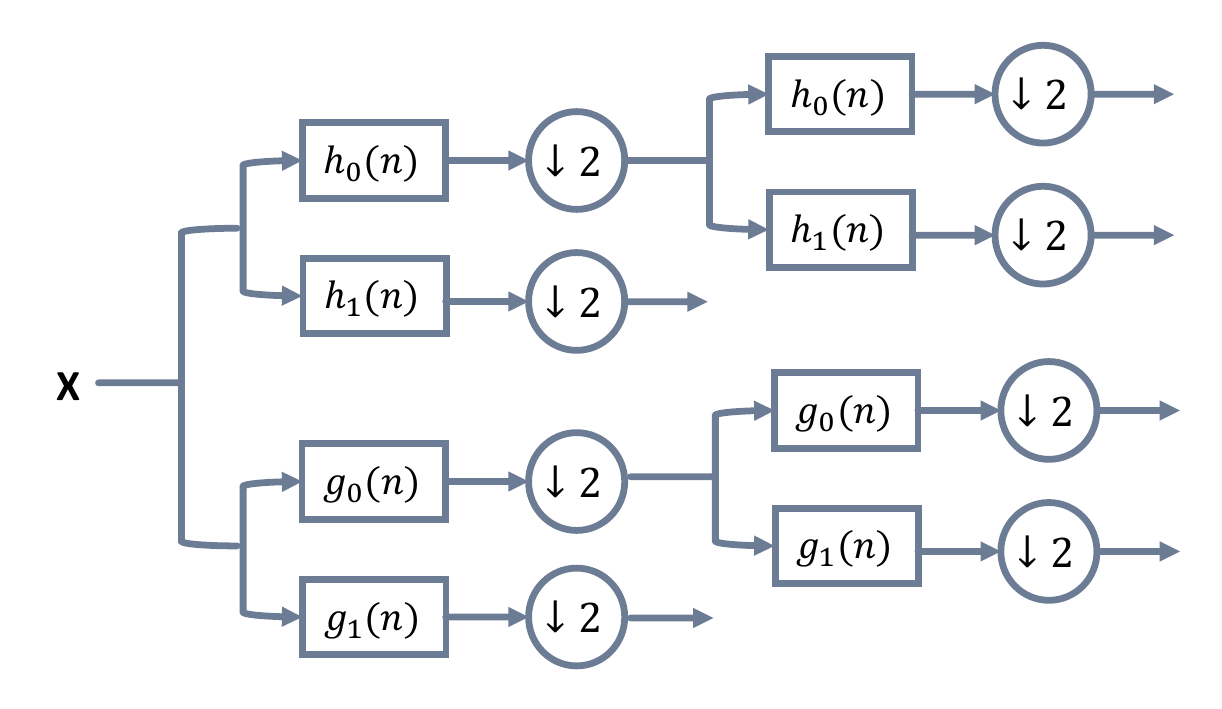}
    \caption{\textbf{Analysis Filter Bank}, for the dual tree complex wavelet transform.}
    \label{fig:dtcwt_filter_bank}
\vspace{-1em}
\end{figure}

Built upon plane-based 4D volume decomposition and drawing inspiration from the dual-tree complex wavelet transform, we introduce a direction-aware representation. This innovative approach enables the modeling of representations from six different directions. In contrast to the prevalent use of 2D discrete wavelet transforms (DWT), the dual tree complex wavelet transform (DTCWT) \cite{selesnick2005dual} employs two complex wavelets as illustrated in Figure \ref{fig:dtcwt_filter_bank}.
Given $h=[h_0,h_1]$ and $g=[g_0,g_1]$ low/high pass filter pairs for upper (real) and lower (imaginary) filter banks, the low-pass and high-pass complex wavelet transforms in DTCWT are denoted as $\phi(x)=\phi_h(x)+j\phi_g(x)$ and $\psi(x)=\psi_h(x)+j\psi_g(x)$.
Consequently, applying low- and high-pass complex wavelet transforms to rows and columns of a 2D grid yields wavelet coefficients $\phi(x)\psi(y)$, $\psi(x)\phi(y)$ and $\psi(x)\psi(y)$. Due to filter design, the upper (real) and lower (imaginary) filter satisfy the Hilbert transform, denoted as $\psi_g(x)\approx\mathcal{H}(\psi_h(x))$. Finally, three additional wavelet coefficients,  $\phi(x)\overline{\psi(y)}$, $\psi(x)\overline{\phi(y)}$ and $\psi(x)\overline{\psi(y)}$, can be obtained, where $\overline{\phi}$ and $\overline{\psi}$ represent the complex conjugate of $\phi$ and $\psi$.  From these 2D wavelet coefficients, we derive six direction-aware real and imaginary wavelet coefficients, each with the same set of six directions. Compared to 2D DWT, the six wavelet coefficients align along specific directions, eliminating the checkerboard effect, with more results in the supplementary material.

Exploiting the properties of DTCWT, we aim for the plane-based representation $M_r^{AB}\in\mathbb{R}^{m,n}$ of the 4D volume to possess direction-aware capabilities as illustrated in the top section of Figure \ref{fig:dtcwt_filter_bank}. Here, $m$ and $n$ denote the resolution of the 2D plane-based representation. To imbue each 2D plane-based representation with direction-aware capabilities, we introduce twelve learned wavelet coefficients—six for the real part and six for the imaginary part—denoted as $\mathbf{R}\{\mathcal{W}_i^{AB}\}_{i=1}^6 \in\mathbb{R}^{m/2^l,n/2^l}$ and $\mathbf{I}\{\mathcal{W}_i^{AB}\}_{i=1}^6 \in\mathbb{R}^{m/2^l,n/2^l}$, respectively. Additionally, a learned approximation 
coefficient is defined as $\mathcal{W}_a^{AB}\in\mathbb{R}^{m/2^{l-1},n/2^{l-1}}$, with 
$l$ the DTCWT transformation level. Consequently, a specific plane-based representation can be expressed as:
\begin{small}
\begin{equation}
    M_r^{AB} = IDTCWT([W_{a,r}^{AB},\mathbf{R}\{\mathcal{W}_{i,r}^{AB}\}_{i=1}^6,\mathbf{I}\{\mathcal{W}_{i,r}^{AB}\}_{i=1}^6])
\end{equation}
\end{small}
Importantly, our 
representation is not only applicable for modeling dynamic 3D scenes but is also well-suited for static 3D scenes, following a TensorRF-like \cite{chen2022tensorf} decomposition:
\begin{small}
\begin{equation}
\begin{split}
    D=&\sum_{r=1}^{R_1}{M_r^{XY}\circ v_r^{Z}\circ v_r^1} +\sum_{r=1}^{R_2}{M_r^{XZ}\circ v_r^{Y}\circ v_r^2} \\
    &+ \sum_{r=1}^{R_3}{M_r^{YZ}\circ v_r^{X}\circ v_r^3}
\end{split}
\end{equation} 
\end{small}
In this formulation, a plane-based representation $M_r^{AB}\in \mathbb{R}^{AB}$ and a vector-based representation $v_r^C\in\mathbb{R}^C$ are employed to model a 3D volume. For static scenes, our direction-aware representations also could be applied to represent the plane-based representations.

\begin{figure}[t]
    \centering
    \includegraphics[width= 0.5\textwidth, height = 0.4\textwidth]{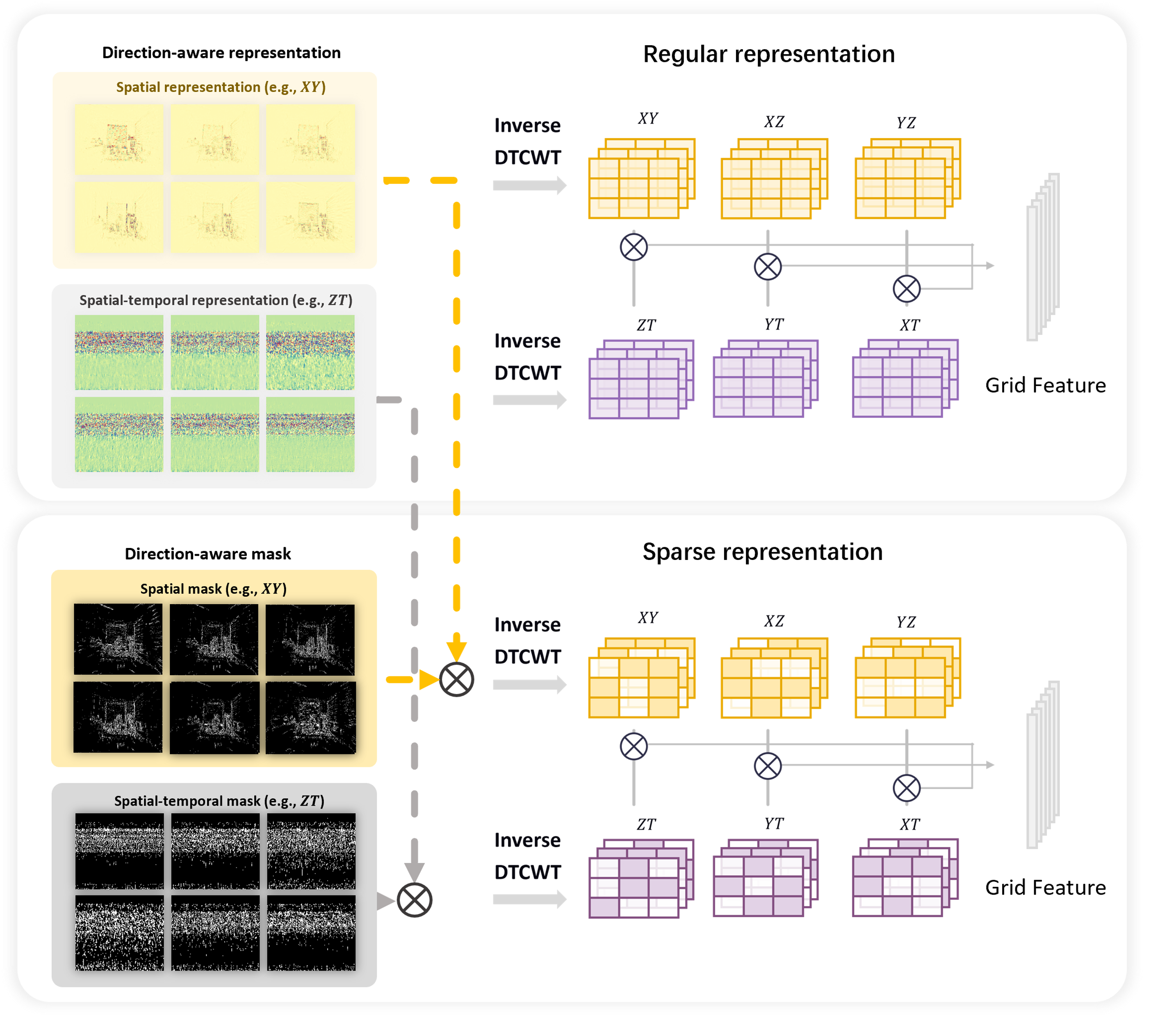} 
    \caption{\textbf{DaRePlane and DaRePlane-S Overview.} \textbf{Top:} The regular DaRePlane architecture comprises an approximation and 12 direction-aware coefficient maps for both spatial (\eg, $XY$) and spatial-temporal (\eg, $ZT$) plane-based representation. To compute the features of points in space-time, it multiplies feature vectors extracted from paired planes (\eg, $XY$ and $ZT$). \textbf{Bottom:} The trainable mask is combined with the top architecture to create DaRePlane-S. Each direction-aware representation and the approximation representation are assigned their own sparse masks. The sparse representation undergoes an inverse dual tree complex wavelet transform to obtain plane-based spatial and spatial-temporal representations.}
    \label{fig:dareplane_arch}
\end{figure}

\subsection{Sparse Representation and Model Compression}
In contrast to the classical 2D discrete wavelet transform (DWT), our direction-aware representation excels in modeling dynamic 3D scenes. However, it is worth noting that a single-level dual tree complex wavelet transform (DTCWT) necessitates six real direction-aware wavelet coefficients and six imaginary direction-aware wavelet coefficients to impart directional information to the plane-based representation. In contrast, a single-level 2D DWT only has three real wavelet coefficients, albeit with inherent direction ambiguity. To enhance the storage efficiency of our solution, we employ learned masks \cite{rho2023masked} for each directional wavelet coefficient, selectively masking out less important features.

As illustrated in the bottom section of Figure \ref{fig:dareplane_arch}, to address the $2^d$ redundancies, where $d=2$ for the 2D DTCWT transform, we employ learned masks $\mathbf{R}\{\mathcal{M}_i^{AB}\}_{i=1}^6 \in\mathbb{R}^{m/2^l,n/2^l}$, $\mathbf{I}\{\mathcal{M}_i^{AB}\}_{i=1}^6 \in\mathbb{R}^{m/2^l,n/2^l}$ and $\mathcal{M}_a^{AB}\in\mathbb{R}^{m/2^{l-1},n/2^{l-1}}$ for the six real wavelet coefficients, six imaginary wavelet coefficients and the approximation coefficients, respectively. The masked wavelet coefficients can be denoted as:
\begin{equation}
    \widehat{\mathcal{W}}^{AB}=\sg\Big(\big(\mathbf{H}(\mathcal{M}^{AB})-\sigmoid(\mathcal{M}^{AB})\big)\odot\mathcal{W}^{AB}\Big),
\end{equation}
with $\big\{\mathbf{R}\{\mathcal{M}_i^{AB}\}_{i=1}^6,\mathbf{I}\{\mathcal{M}_i^{AB}\}_{i=1}^6,\mathcal{M}_a^{AB}\big\}\in\mathcal{M}^{AB}$ and $\big\{\mathbf{R}\{\mathcal{W}_i^{AB}\}_{i=1}^6,\mathbf{I}\{\mathcal{W}_i^{AB}\}_{i=1}^6,\mathcal{W}_a^{AB}\big\}\in\mathcal{W}^{AB}$. The functions $\sg$, $\mathbf{H}$ and $\sigmoid$ represent the stop-gradient operator, Heaviside step and element-wise sigmoid function, respectively. The masked plane-based representation is obtained from the masked wavelet coefficients through the equation:
\begin{small}
\begin{equation}
    \widehat{M}_r^{AB} = IDTCWT([\widehat{W}_{a,r}^{AB},\mathbf{R}\{\widehat{\mathcal{W}}_{i,r}^{AB}\}_{i=1}^6,\mathbf{I}\{\widehat{\mathcal{W}}_{i,r}^{AB}\}_{i=1}^6])
\end{equation}
\end{small}
To encourage sparsity in the generated masks, we introduce an additional loss term $\mathcal{L}_m$, defined as the sum of all masks. We employ $\lambda_m$ as the weight of $\mathcal{L}_m$ to control the sparsity of the representation.

Following the removal of unnecessary representations through masking, we adopt a compression strategy akin to the one employed in masked wavelet NeRF \cite{rho2023masked}, originally designed for static scenes, to compress the sparse representation and masks that identify non-zero elements. The process involves converting the binary mask values to 8-bit unsigned integers and subsequently applying run-length encoding (RLE). Finally, the Huffman encoding algorithm is employed on the RLE-encoded streams to efficiently map values with a high probability to shorter bits.


\subsection{Optimization}
We leverage our proposed direction-aware representation to effectively represent 3D dynamic scenes. The model is then optimized through a photometric loss function, which measures the difference between rendered images and target images. Additionally, we also add regularization items to reduce the artifacts and utilize the mask loss to control the sparsity of the DaRePlane. The overall loss is expressed as:
\begin{equation}
    \mathcal{L} = \lambda_{photo}\mathcal{L}_{photo} + \lambda_{reg}\mathcal{L}_{reg} + \lambda_m\mathcal{L}_m,
\end{equation}
with $\mathcal{L}_{photo}$, $\lambda_{photo}$, $\mathcal{L}_{reg}$, $\lambda_{reg}$ and $\mathcal{L}_{m}$, $\lambda_{m}$ the photometric loss, regularization loss and mask loss with respective weights. For both the NeRF and GS setting, we utilize the total variational (TV) loss as regularization item.\\
\subsubsection{NeRF} 
In the NeRF framework, for a given point $(x,y,z,t)$, its opacity and appearance features are represented by DaRePlane. The final color is obtained through a small multi-layer perceptron (MLP), which takes the appearance feature and view direction as inputs. Using the point's opacities and colors, images are generated through volumetric rendering.\\
\noindent
\textbf{Photometric Loss.} For the photometric loss $\mathcal{L}_{photo}$, we utilize the mean square error (MSE) as the loss function. \\
\noindent
\textbf{Training Strategy}. We employ the same coarse-to-fine training strategy as in  \cite{cao2023hexplane,chen2022tensorf,yu2021plenoctrees}, where the resolution of grids progressively increases during training. This strategy not only accelerates the training process but also imparts an implicit regularization on nearby grids.\\
\noindent
\textbf{Emptiness Voxel}. We maintain a small 3D voxel representation that indicates the emptiness of specific regions in the scene, allowing us to skip points located in empty regions. Given the typically large number of empty regions, this strategy significantly aids in acceleration. To generate this voxel, we evaluate the opacities of points across different time steps and aggregate them into a single voxel by retaining the maximum opacities. While preserving multiple voxels for distinct time intervals could potentially enhance efficiency, for the sake of simplicity, we opt to keep only one voxel.

\subsubsection{GS} In the GS framework, we obtain the spatial-temporal representation, denoted as $\mathbf{f}$, from DaRePlane, we use four tiny MLPs, denoted as $\mathbf{F} = \{\mathbf{F}_{\mathbf{\mu}},\mathbf{F}_{\mathbf{R}},\mathbf{F}_{\mathbf{S}},\mathbf{F}_{\mathbf{o}}\}$, to predict the time-variant deformation of position, rotation, scaling, and opacity of Gaussians, respectively. 
With the deformation of position $\Delta \bm{\mu} = \mathbf{F_{\mu}(\mathbf{f})}$, rotation $\Delta \mathbf{R} = \mathbf{F_{R}(\mathbf{f})}$, scaling $\Delta \mathbf{S} = \mathbf{F_{S}(\mathbf{f})}$, opacity $\Delta \mathbf{o} = \mathbf{F_{o}(\mathbf{f})}$, 
the time-variant deformed Gaussians $\mathbf{G_t}$ 
at time \(t\) can be expressed as：
\begin{equation}
    \mathbf{G}_t = \mathbf{G}_\mathbf{0} + \Delta \mathbf{G} = (\bm{\mu}+\Delta\bm{\mu}, \mathbf{R} + \Delta\mathbf{R}, \mathbf{S} + \Delta\mathbf{S},\mathbf{o} + \Delta\mathbf{o})
\end{equation}
\textbf{Photometric Loss.} The photometric loss for the Gaussian Splatting setting consist of two main components: 1) color losses, and 2) depth loss as shown below:
\begin{equation}
    \mathcal{L}_{color} = \sum_{x\in \mathcal{I}}\left \| \mathbf{M}(\mathbf{x})(\hat{\mathbf{C}}(\mathbf{x})-\mathbf{C}(\mathbf{x})) \right \|_1
\end{equation}
\begin{small}
\begin{equation}
    \mathcal{L}_{depth} = 1 - Cov(\mathbf{M}\odot \hat{\mathbf{D}}, \mathbf{M}\odot \mathbf{D})/ \sqrt{Var(\mathbf{M} \odot \hat{\mathbf{D}})Var(\mathbf{M}\odot\mathbf{D})}
    \label{eq:gs_depth_loss}
\end{equation}
\end{small}
where $\mathbf{M}$, $\{\hat{\mathbf{C}},\hat{\mathbf{D}}\}$, $\{\mathbf{C}, \mathbf{D}\}$, and $\mathcal{I}$ are binary tool masks, predicted colors and depths, real colors and depths, and 2D coordinate space, respectively. $Cov$ and $Var$ operations in \ref{eq:gs_depth_loss} represent the covariance and variances of the prediction and ground truth, respectively. This is equivalent to using the Pearson Correlation Coefficients (PCC) as loss.

\section{Experiments}

We first demonstrate the capabilities of our DaRePlane system on real-world dynamic and static scenes within the NeRF framework. We conduct a comprehensive comparison with existing methods and investigate the advantages of DaRePlane through extensive ablation studies. These studies showcase DaRePlane's robustness and effectiveness in handling both dynamic and static scenes.

Next, we demonstrate that DaRePlane is suitable for entirely different scenarios compared to regular scenes and can transfer flexibly to different rendering systems, such as Gaussian Splatting. We evaluate DaReNeRF and DaReGS across different types of surgical scenes to highlight its versatility and effectiveness in these specialized applications.
\subsection{Novel View Synthesis for Regular Dynamic Scenes}
For dynamic scenes, we employ two distinct datasets with varying settings. 
Each dataset presents its own challenges, effectively addressed by our direction-aware representation.

\noindent
\textbf{Plenoptic Video Dataset} \cite{li2022neural} is a real-world dataset captured by a multi-view camera system using 21 GoPro cameras at a resolution of $2028 \times 2704$ and a frame rate of 30 frames per second. Each scene consists of 19 synchronized, 10-second videos, with 18 videos designated for training and one for evaluation. This dataset serves as an ideal testbed to assess the representation ability, featuring complex and challenging dynamic content, including highly specular, translucent, and transparent objects, topology changes, moving self-casting shadows, fire flames, strong view-dependent effects for moving objects, and more.

For a fair and direct comparison, we adhere to the same training and evaluation protocols as DyNeRF \cite{li2022neural}. Our model is trained on a single A100 GPU, utilizing a batch size of 4,096. We adopt identical importance sampling strategies and hierarchical training techniques as DyNeRF, employing a spatial grid size of 512 and a temporal grid size of 300. The scene is placed under the normalized device coordinates (NDC) setting, consistent with the approach outlined in \cite{mildenhall2021nerf}.

\begin{figure*}[t]
    \centering
  

    \includegraphics[width=1.\linewidth]{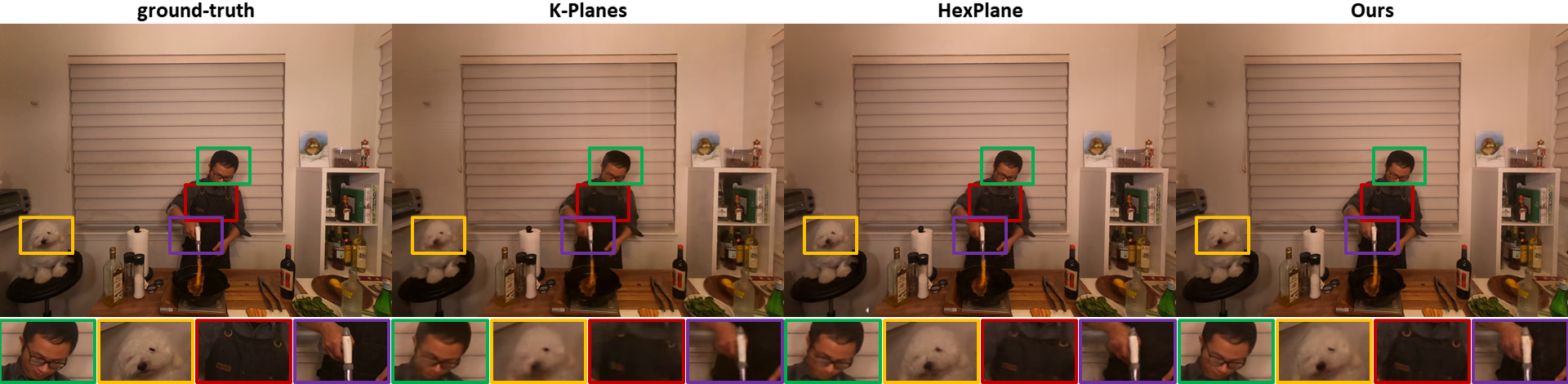}
    \caption{
    \textbf{Visual Comparison on Dynamic Scenes (Plenoptic Data).}
    K-Planes and HexPlane are concurrent decomposition-based methods. 
    As shown in the four zoomed-in patches, our method better reconstruct fine details and captures motion.
    Please refer to the supplementary material to see the figure animated.}
    
    \label{fig:Visual_results_plenoptic}
\vspace{-1em}
\end{figure*}

Quantitative compression results with state-of-the-art methods are presented in Table \ref{tab:plenoptic_table}. We utilize measurement metrics PSNR, structure dissimilarity index measure (DSSIM) \cite{sara2019image}, and perception quality measure LPIPS \cite{zhang2018unreasonable} to conduct a comprehensive evaluation. As demonstrated in Table \ref{tab:plenoptic_table}, leveraging the proposed direction-aware representation, both regular and sparse DaReNeRF achieve promising results compared to the most recent state-of-the-art, with analogous training time. This more ideal trade-off between performance and computational requirements, compared to prior art, is also illustrated in Figure \ref{fig:figure_1}.b, computed over Plenoptic data.
Figure \ref{fig:Visual_results_plenoptic} presents some novel-view  results on the Plenoptic dataset. Four small patches, each containing detailed texture information, are selected for comparison. DaReNeRF, equipped with the proposed direction-aware representation, excels in reconstructing moving objects (\eg, dog and firing gun) and capturing better texture details (\eg, hair and metal rings on the apron).
\begin{table*}[ht]
    \caption{\textbf{Quantitative Comparison on Plenoptic Video Data.} We present results on synthesis quality and training time (measured in GPU hours).
    Following prior art, we provide both scene-specific performance ({\footnotesize\texttt{flame-salmon}} scene) and mean performance across all cases from their original papers.}
    \centering
    \setlength{\tabcolsep}{3.5mm}
    \resizebox{\linewidth}{!}{
    
    \begin{tabular}{rccccccc}
        \toprule
         & Model&Steps&PSNR$\uparrow$&D-SSIM$\downarrow$&LPIPS$\downarrow$&Training Time$\downarrow$&Model Size (MB) $\downarrow$\\
         \midrule
         \parbox[t]{0mm}{\multirow{8}{*}{\rotatebox[origin=c]{90}{{\scriptsize\texttt{flame-salmon}} scene}}} &
         Neural Volumes \cite{lombardi2019neural}&-&22.800&0.062&0.295&-\\
         & LLFF \cite{mildenhall2019local}&-&23.239&0.076&0.235&-&-\\
         & NeRF-T \cite{li2022neural}&-&28.449&0.023&0.100&-&-\\
         & DyNeRF \cite{li2022neural}&650k&29.581&0.020&0.099&1,344h&\textbf{28}\\
         & HexPlane \cite{cao2023hexplane}&650k&29.470&0.018&\textbf{0.078}&12h& 252\\
         & HexPlane \cite{cao2023hexplane}&100k&29.263&0.020&0.097&\textbf{2h}& 252\\
         & DaReNeRF-S&100k&\underline{30.224}&\underline{0.015}&0.089&5h& \underline{244}\\
         & DaReNeRF&100k&\textbf{30.441}&\textbf{0.012}&\underline{0.084}&\underline{4.5h}& 1,210 \\
         \midrule
         \parbox[t]{0mm}{\multirow{11}{*}{\rotatebox[origin=c]{90}{all scenes (average)}}} &
         NeRFPlayer \cite{song2023nerfplayer}&-&30.690&0.034&0.111&6h&-\\
         & HyperReel \cite{attal2023hyperreel}&-&31.100&0.036&0.096&9h&-\\
         & HexPlane \cite{cao2023hexplane}&650k&31.705&\underline{0.014}&\underline{0.075}&12h& 252\\
         & HexPlane \cite{cao2023hexplane}&100k&31.569&0.016&0.089&\underline{2h}& 252\\
         & K-Planes-explicit \cite{fridovich2023k}&120k&30.880&-&-&3.7h& 580\\
         & K-Planes-hybrid&90k&31.630&-&-&1.8h& 310\\
         & Mix Voxels-L \cite{wang2023mixed}&25k&31.340&0.019&0.096&\textbf{1.3h}& 500\\
         & Mix Voxels-X \cite{wang2023mixed}&50k&31.730&0.015&\textbf{0.064}&5h& 500\\
         & 4D-GS \cite{wu20234d}&-&31.020&-&0.150&\underline{2h}&\textbf{145}\\
         & DaReNeRF-S&100k&\underline{32.102}&\underline{0.014}&0.087&5h& \underline{244}\\
         & DaReNeRF&100k&\textbf{32.258}&\textbf{0.012}&0.084&4.5h& 1,210\\
         \bottomrule
         
    \end{tabular}
    }

    \label{tab:plenoptic_table}
\vspace{-1em}
\end{table*}

\noindent
\textbf{D-NeRF Dataset} \cite{pumarola2021d} is a monocular video dataset with $360^\circ$ observations of synthetic objects. Dynamic 3D reconstruction from monocular video poses challenges as only one observation is available for each timestamp. State-of-the-art methods for monocular video typically incorporate a deformation field. 
However, the underlying assumption is that the scenes undergo no topological changes,
making them less effective for real-world cases (\eg, Plenoptic dataset). 
Table \ref{tab:dnerf_table} reports the rendering quality of different methods with and without deformation fields on the D-NeRF data, 
DaReNeRF outperforms all non-deformation methods, as well as some deformation methods, \eg D-NeRF and TiNeuVox-S \cite{fang2022fast}. 
The superiority of our solution on topologically-changing scenes is further highlighted in appendix.

\begin{table}[t]
    \centering
    \caption{
    \textbf{Quantitative Study on D-NeRF Data.}
    Without 
    the topological constraints of using deformation fields, 
    DaReNeRF outperforms even some deformation-based methods.}
    \vspace{-.5em}
    \resizebox{\linewidth}{!}{
    \begin{tabular}{ccccc}
    \toprule
         Model&Deform.&PSNR$\uparrow$&SSIM$\uparrow$&LPIPS$\downarrow$  \\
         \midrule
         D-NeRF \cite{pumarola2021d}&\checkmark&30.50&0.95&0.07\\
         TiNeuVox-S \cite{fang2022fast}&\checkmark&30.75&0.96&0.07\\
         TiNeuVox-B \cite{fang2022fast}&\checkmark&\underline{32.67}&\underline{0.97}&\underline{0.04}\\
         4D-GS \cite{wu20234d}&\checkmark&\textbf{33.30}&\textbf{0.98}&\textbf{0.03}\\
         \midrule
         T-NeRF \cite{pumarola2021d}&-&29.51&\underline{0.95}&0.08\\
         HexPlane \cite{cao2023hexplane}&-&31.04&\textbf{0.97}&\underline{0.04}\\
         K-Planes \cite{fridovich2023k}&-&31.05&\textbf{0.97}&-\\
         DaReNeRF-S&-&\underline{31.82}&\textbf{0.97}&\textbf{0.03}\\
         DaReNeRF&-&\textbf{31.95}&\textbf{0.97}&\textbf{0.03}\\
         \bottomrule
    \end{tabular}
    }
    \label{tab:dnerf_table}
\vspace{-1.5em}
\end{table}

\subsection{Novel View Synthesis of Regular Static Scenes}
For static scenes, we test our proposed direction-aware representation on NeRF synthetic \cite{mildenhall2021nerf}, Neural Sparse Voxel Fields (NSVF) \cite{liu2020neural} and LLFF \cite{mildenhall2019local} datasets. We use TensoRF-192 as baseline and apply our proposed representation. We report the performance on these three datasets in Tables \ref{tab:nerf_table}, \ref{tab:NVSF_table}, and \ref{tab:LLFF_table} respectively.

\begin{table}[t]
    \centering
    \caption{
    \textbf{Quantitative Comparison on NeRF Synth.}, with models designed for different bit-precisions ($^\ast$ denotes a model quantized post-training;  numbers in brackets denote grid resolutions).}
    \vspace{-.75em}
    \label{tab:nerf_table}
    \setlength{\tabcolsep}{3.5mm}
    \resizebox{\linewidth}{!}{
    \begin{tabular}{cccc}
        \toprule
         Precision&Method& Size (MB) & PSNR $\uparrow$\\
        \midrule
         32-bit&KiloNeRF \cite{reiser2021kilonerf}& $\leq$ 100& 31.00\\
         32-bit&CCNeRF (CP) \cite{tang2022compressible}& 4.4 & 30.55\\
         8-bit$^\ast$&NeRF \cite{mildenhall2021nerf}&1.25&31.52\\
         8-bit&cNeRF \cite{bird20213d}&\textbf{0.70}&30.49\\
         8-bit&PREF \cite{huang2022pref}&9.88&31.56\\
         8-bit$^\ast$&VM-192 \cite{chen2022tensorf} & 17.93&\textbf{32.91}\\
         8-bit$^\ast$&VM-192 (300) + DWT \cite{rho2023masked}&\underline{0.83}&31.95\\
         \midrule
         8-bit$^\ast$&VM-192 (300) + Ours &8.91&\underline{32.42}\\
         \bottomrule
         
    \end{tabular}
    }
\vspace{-.75em}
\end{table}

\begin{table}[t]
    \centering
    \caption{
    \textbf{Quantitative Comparison on NSVF} (static scenes).}
    \vspace{-.5em}
    \setlength{\tabcolsep}{3.5mm}
    \resizebox{\linewidth}{!}{
    \begin{tabular}{cccc}
    \toprule
         Bit Precision&Model & Size (MB) & PSNR $\uparrow$  \\
         \midrule
         32-bit&KiloNeRF \cite{reiser2021kilonerf}& $\leq$ 100& 33.37\\ 
         8-bit$^\ast$&VM-192 \cite{song2022pref} & 17.77 &\underline{36.11}\\
         8-bit$^\ast$&VM-48  \cite{chen2022tensorf}& 4.53 &34.95\\
         8-bit$^\ast$&CP-384 \cite{chen2022tensorf}& \textbf{0.72} &33.92\\
         8-bit$^\ast$&VM-192 (300) + DWT \cite{rho2023masked}&\underline{0.87}&34.67\\
         \midrule
         8-bit$^\ast$&VM-192 (300) + Ours &8.98&\textbf{36.24}\\
         \bottomrule
    \end{tabular}
    }
    \label{tab:NVSF_table}
\vspace{-.75em}
\end{table}

\begin{table}[t]
    \centering
    \caption{
    \textbf{Quantitative Comparison on LLFF} (static scenes).}
    \vspace{-.75em}
    \setlength{\tabcolsep}{3.5mm}
    \resizebox{\linewidth}{!}{
    \begin{tabular}{cccc}
    \toprule
         Bit Precision&Model & Size(MB) & PSNR $\uparrow$  \\
         \midrule
         8-bit&cNeRF \cite{bird20213d}& \underline{0.96}&26.15\\
         8-bit$^\ast$&PREF \cite{song2022pref} &9.34&24.50\\
         8-bit$^\ast$&VM-96  \cite{chen2022tensorf}&44.72&\textbf{26.66}\\
         8-bit$^\ast$&VM-48  \cite{chen2022tensorf}&22.40&26.46\\
         8-bit$^\ast$&CP-384 \cite{chen2022tensorf}&\textbf{0.64}&25.51\\
         8-bit$^\ast$&VM-96 (640) + DWT \cite{rho2023masked}&1.34&25.88\\
         \midrule
         8-bit$^\ast$&VM-96 (640) + Ours&13.67&\underline{26.48}\\
         \bottomrule
    \end{tabular}
    }
    
    \label{tab:LLFF_table}
\vspace{-1.5em}
\end{table}

    Across these three static datasets, our direction-aware representation outperforms most compression-based NeRF models with model sizes ranging from 8 to 14MB. While our method's model size is larger than DWT-based solutions, it achieves comparable sparsity. 
For instance, with $\lambda_m=2.5\times 10^{-11}$, its \textit{sparsity} reaches 94$\%$, closely aligned with the 97$\%$ reported in the masked wavelet NeRF \cite{rho2023masked} paper. Notably, with similar sparsity, our direction-aware method exhibits PSNR improvements of 0.47, 1.57, and 0.60 over DWT-based methods on the three static datasets.

\begin{figure}[t]
    \centering
    \includegraphics[width=1.\linewidth]{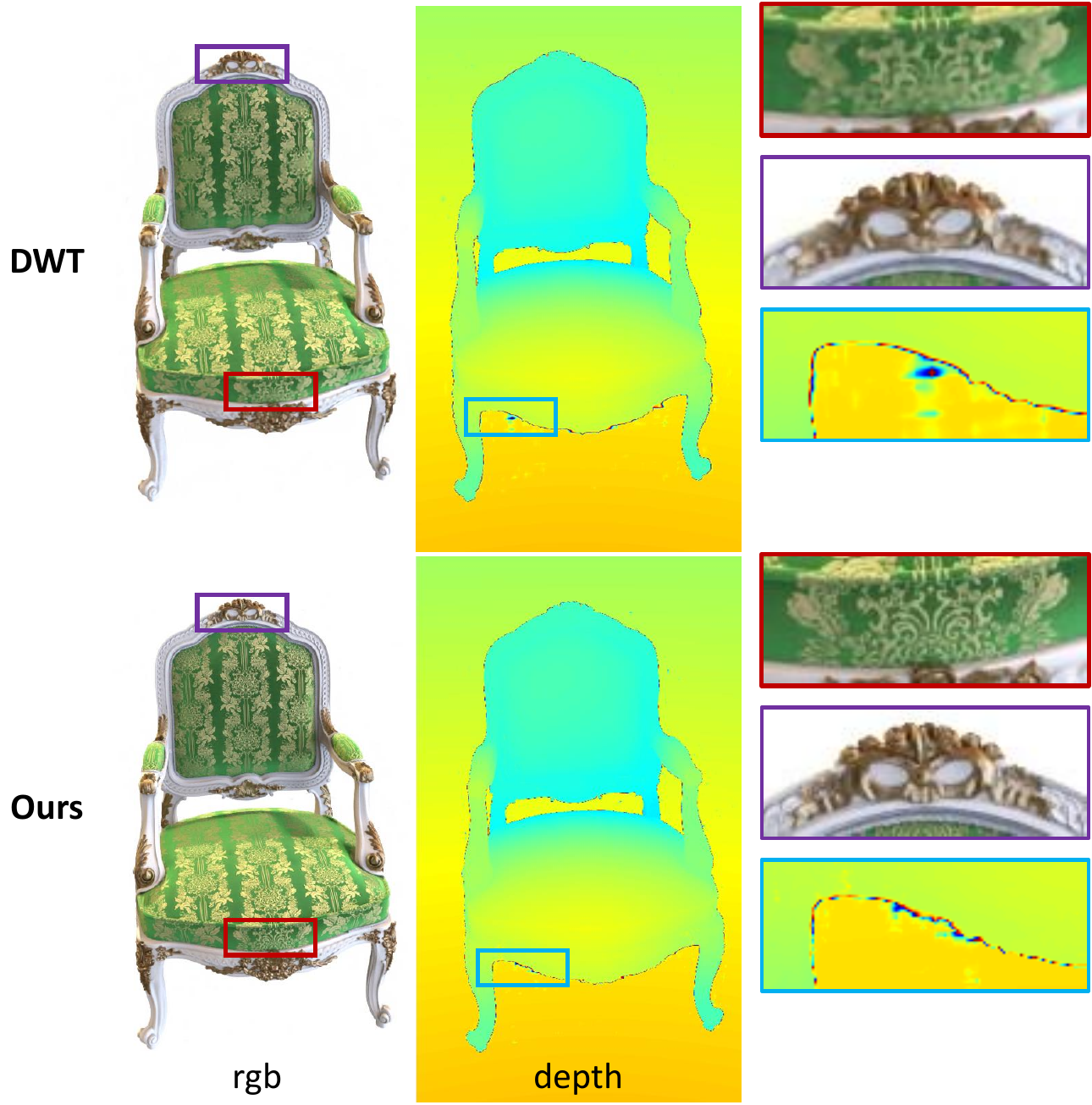}
    \caption{
    \textbf{Visual Comparison of Static Scenes on NSVF Data.}
    Two representative patches are selected for closer inspection. Our method, free from the DWT limitations of shift variance and direction ambiguity, achieves superior texture reconstruction performance.}
    \label{fig:static_comp}
\vspace{-1.5em}
\end{figure}

Figure \ref{fig:static_comp} highlights the qualitative differences between DWT-based solutions and our proposed direction-aware method. In static scenes, our solution excels in reconstructing texture details compared to DWT representation, which is less sensitive to lines and edges patterns due to shift variance and direction ambiguity.

\subsection{Novel View Synthesis of Dynamic Surgical Scenes}
For dynamic surgical scene reconstruction, we employ four more distinct datasets with various types of surgical setting. Each dataset has different camera settings and its own challenges.

\noindent
\textbf{EndoNeRF Dataset \cite{wang2022neural}.} The data was obtained from DaVinci robotic prostatectomy videos. Six clips, totaling 807 frames, were extracted from these videos, with each clip lasting 4 to 8 seconds at 15 \textit{fps}. The footage is captured from stereo cameras at a single viewpoint and encompasses challenging scenes with non-rigid deformation and tool occlusion. Due to the privacy of the surgical data, only two clips from this dataset are publicly available: \texttt{cutting tissues} and \texttt{pulling}.

\begin{table}[t]
    \centering
    \caption{\textbf{Quantitative Comparison on EndoNeRF Dataset} (Surgical Scene).}
    \resizebox{\linewidth}{!}{
    \begin{tabular}{rccccc}
    \toprule
       &Method& PSNR $\uparrow$& SSIM $\uparrow$& LPIPS $\downarrow$& Training time $\downarrow$ \\
       \midrule
       \parbox[t]{0mm}{\multirow{4}{*}{\rotatebox[origin=c]{90}{{\scriptsize\texttt{NeRF}}}}}& EndoNeRF \cite{wang2022neural} & 36.062& 0.933& 0.089& 12.0 hours \\
       &EndoSurf \cite{zha2023endosurf}& \underline{36.529}& \textbf{0.954}& \textbf{0.074}& 8.5 hours\\
       &LerPlane \cite{yang2023neural}& 34.988& 0.926& 0.080& \textbf{3.5} mins\\
       &DaReNeRF &\textbf{36.685}& \underline{0.947}& \underline{0.076}& \underline{4.0} mins\\
       \midrule
       \parbox[t]{0mm}{\multirow{2}{*}{\rotatebox[origin=c]{90}{{\scriptsize\texttt{GS}}}}}&EndoGaussian \cite{liu2024endogaussian} & \underline{37.553}& \underline{0.959}& \underline{0.059}&\textbf{2.0} mins\\
       &DaReNeGS &\textbf{38.348}&\textbf{0.966}&\textbf{0.040}&\underline{3.5} mins\\
       \bottomrule

    \end{tabular}
    }
    
    \label{endonerf_dataset}
\end{table}

\noindent
\textbf{Hamlyn Dataset \cite{mountney2010three, stoyanov2005soft}.} The Hamlyn dataset includes both phantom heart and in-vivo sequences captured during da Vinci surgical robot procedures. The rectified images, stereo depth, and camera calibration information are sourced from \cite{recasens2021endo}. To generate instrument masks, we use the Vision Foundation Model, Segment Anything Model \cite{kirillov2023segment}, which enables the segmentation of surgical instruments. The Hamlyn dataset presents a rigorous evaluation scenario as it contains sequences depicting intracorporeal scenes with various challenges, such as weak textures, deformations, reflections, surgical tool occlusion, and illumination variations. Similar to previous work \cite{yang2024efficient}, we select seven specific sequences from the Hamlyn dataset (sequences: {\texttt{rectified01}, \texttt{rectified06}, \texttt{rectified08}, and \texttt{rectified09}), each comprising 301 frames with a resolution of 480 × 640. These sequences span approximately 10 seconds and feature scenarios involving surgical tool occlusion and extensive tissue exposure. The frames in each sequence are divided into two sets: 151 frames for training and the remainder for evaluation.

\begin{table}[t]
    \centering
    \caption{\textbf{Quantitative Comparison on Hamlyn} (Surgical Scene).}
    \resizebox{\linewidth}{!}{
    \begin{tabular}{ccccc}
    \toprule
       Method& PSNR $\uparrow$& SSIM $\uparrow$& LPIPS $\downarrow$& Training time $\downarrow$ \\
       \midrule
       E-DSSR \cite{long2021dssr}&18.150&0.640&0.393&13 mins\\
       EndoNeRF \cite{wang2022neural} & 34.879& 0.951& \textbf{0.071}& 12.0 hours \\
       TiNeu Vox-S \cite{fang2022fast} & 35.277& \textbf{0.953}& 0.085& 12 mins\\
       TiNeu Vox-B \cite{fang2022fast} & 33.764& 0.942& 0.146& 90 mins\\
       LerPlane \cite{yang2023neural}& \underline{35.504}& 0.935 &\underline{0.083}&10 mins\\
       ForPlane \cite{yang2024efficient}& 35.301& 0.945& 0.093&\textbf{3.5} mins\\
       DaReNeRF &\textbf{35.641}& \underline{0.952}& 0.085& \underline{4.0} mins\\
       \bottomrule

    \end{tabular}
    }
    
    \label{hamlyn_dataset}
\end{table}

\noindent
\textbf{SCARED Dataset \cite{allan2021stereo}.} The SCARED dataset consists of fresh porcine cadaver abdominal anatomy captured using a da Vinci Xi endoscope and a projector to obtain high-quality depth maps of the scene. We selected five scenes (Sequences: \texttt{dataset1/keyframe1}, \texttt{dataset2/keyframe1}, \texttt{dataset3/keyframe1}, \texttt{dataset6/keyframe1}, and \texttt{dataset7/keyframe1}) from the dataset and split the frame data of each scene into 7:1 training and testing sets, following previous work \cite{liu2024endogaussian}.

\begin{table}[t]
    \centering
    \caption{\textbf{Quantitative Comparison on SCARED} (Surgical Scene).}
    \resizebox{\linewidth}{!}{
    \begin{tabular}{ccccc}
    \toprule
       Method& PSNR $\uparrow$& SSIM $\uparrow$& LPIPS $\downarrow$& Training time $\downarrow$ \\
       \midrule
       EndoNeRF \cite{wang2022neural} & 24.345& 0.768& 0.313& 3.5 hours \\
       EndoSurf \cite{zha2023endosurf} & 25.020& 0.802& 0.356& 5.8 hours\\
       EndoGaussian \cite{liu2024endogaussian} & \underline{27.042}& \underline{0.827}& \underline{0.267}& \textbf{2.2} mins\\
       DaReGS &\textbf{27.500}& \textbf{0.836}& \textbf{0.238}& \underline{3.5} mins\\
       \bottomrule

    \end{tabular}
    }
    
    \label{scared_dataset}
\end{table}

\noindent
\textbf{Cochlear Implant Surgery \cite{lou2023min, lou2023self}.} Unlike the aforementioned data, which were all obtained from the da Vinci Xi endoscope camera, we also selected two of our in-house cochlear implant surgery videos to test our proposed DaRePlane. These surgery videos were captured during real cochlear implant procedures at Vanderbilt University Medical Center (VUMC) and The Medical University of South Carolina (MUSC) using a surgical microscope with I.R.B. approval. 

\begin{table}[t]
    \centering
    \caption{\textbf{Quantitative Comparison on Cochlear Implants Dataset} (Surgical Scene).}
    \resizebox{\linewidth}{!}{
    \begin{tabular}{ccccc}
    \toprule
       Method& PSNR $\uparrow$& SSIM $\uparrow$& LPIPS $\downarrow$& Training time $\downarrow$ \\
       \midrule
       EndoGaussian \cite{liu2024endogaussian} & \underline{34.024}& \underline{0.948}& \underline{0.065}& \textbf{2.2} mins\\
       DaReGS &\textbf{34.386}& \textbf{0.952}& \textbf{0.052}& \underline{3.5} mins\\
       \bottomrule

    \end{tabular}
    }
    
    \label{cochlear_dataset}
\end{table}


Therefore, in Tables \ref{endonerf_dataset}, \ref{hamlyn_dataset}, and \ref{cochlear_dataset}, we present results across various surgical scenarios, including laparoscopy, endoscopy, and microscopy. Our proposed DaRePlane method demonstrates superior performance under both NeRF and Gaussian splatting settings, outperforming previous methods in all surgical scenarios. Specifically, in Table \ref{endonerf_dataset}, our DaReNeRF surpasses all previous NeRF-based methods with only 4 minutes of training time. Additionally, our DaReGS not only outperforms the latest surgical Gaussian Splatting methods but also maintains a similar training time.

In surgical scenarios, explicit representation dynamic scene reconstruction methods such as LerPlane \cite{yang2023neural} and EndoGaussian \cite{liu2024endogaussian} use K-Planes \cite{fridovich2023k} to predict explicit representations and Gaussian deformations, respectively. Similar to regular scenes (based on HexPlane \cite{cao2023hexplane} and TensoRF \cite{chen2022tensorf}), we applied a trainable mask to each plane from the K-Planes method and tested our sparse DaRePlane on the SCARED dataset. From the results in Figure \ref{fig:GS_dareplane}, EndoNeRF and EndoSurf, which are implicit representation-based methods, require significant training time for optimization. However, using explicit representation can achieve up to 100$\times$ acceleration. Additionally, our proposed frequency-based representation combined with trainable masks can save approximately 74\% of memory storage compared to the previous state-of-the-art method, EndoGaussian.

In Figure \ref{fig:surgical_result}, we provide visualization results from three distinct types of surgical datasets, demonstrating the efficacy of our proposed frequency-based representation. This method excels in recovering fine details in dynamic surgical scenes, such as tiny blood vessels and tissue textures. Moreover, our approach adeptly handles complex surgical scenarios, including significant tissue deformation and reflections in endoscopy, as well as transparent tools in microscopy surgery. These capabilities highlight the robustness and versatility of our method in accurately depicting various challenging surgical environments.


    

\begin{figure}
    \centering
    \includegraphics[width=.6\linewidth]{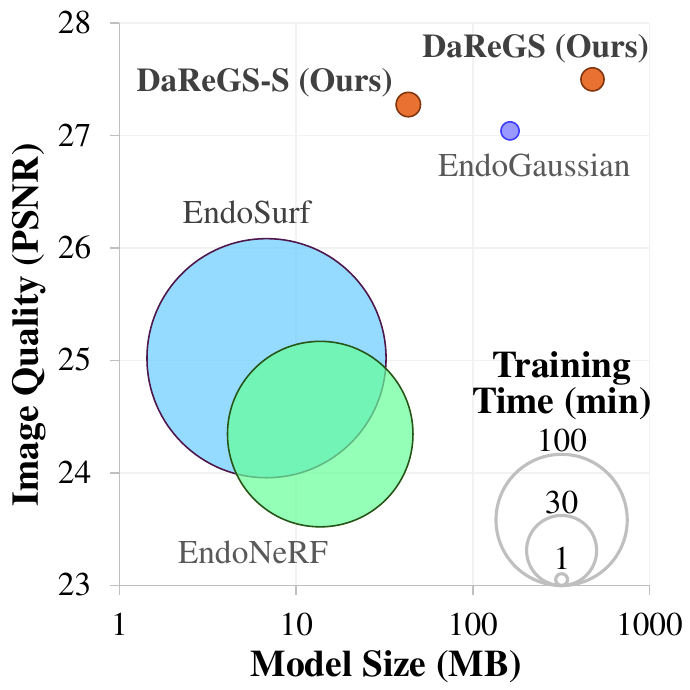}
    \caption{\textbf{Performance of Gaussian Splatting with DaRePlane on 4D scenes.}}
    \label{fig:GS_dareplane}
\vspace{-1em}
\end{figure}

\begin{figure}
    \centering
    \includegraphics[width=1.\linewidth]{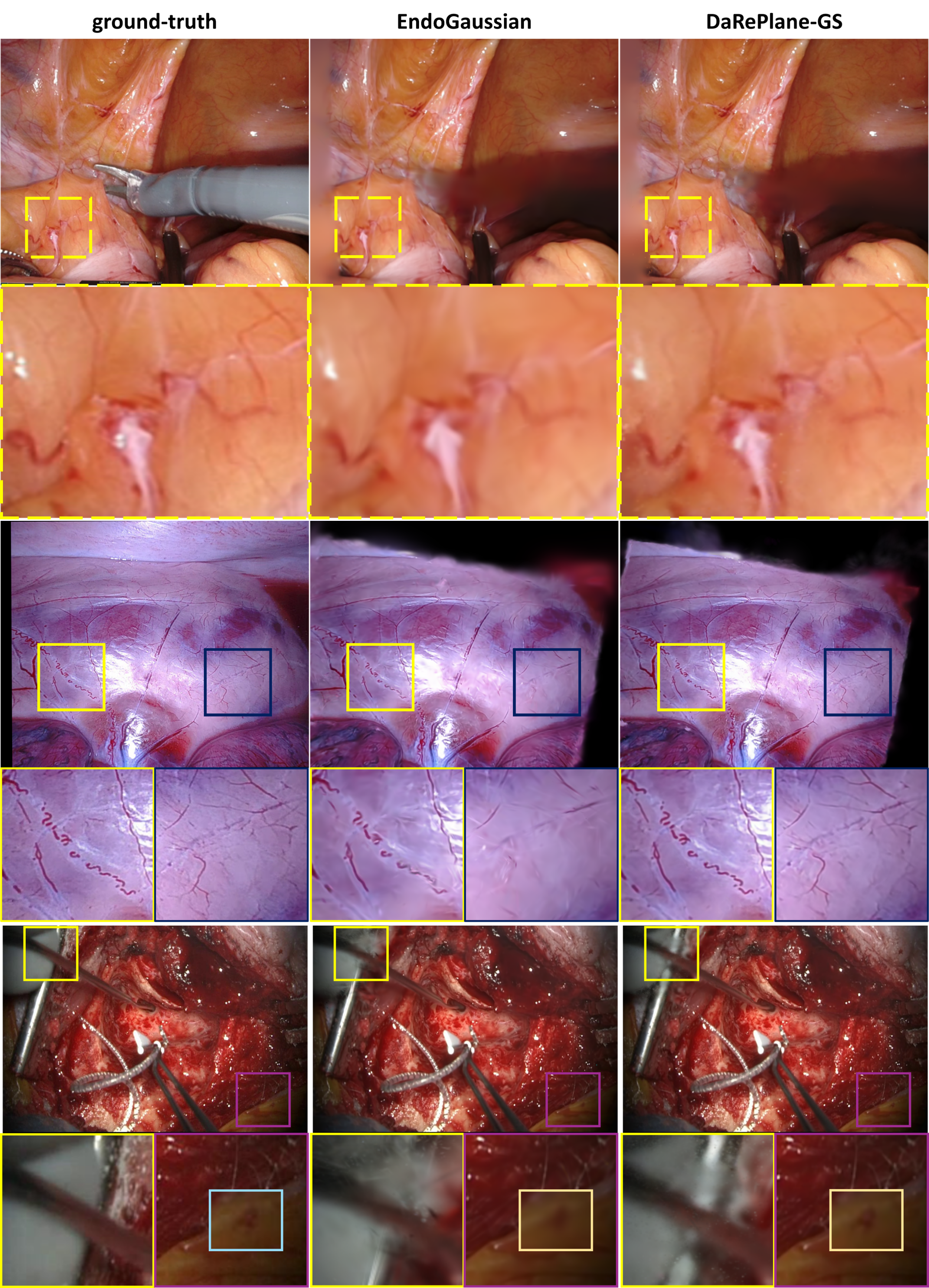}
    \caption{\textbf{Visual Comparison on Surgical Scenes}. From top to bottom, the rows show results from the EndoNeRF, SCARED, and Cochlear Implant datasets. As demonstrated in the zoomed-in patches, our DaRePlane method recovers extremely fine details in the dynamic surgical scenes.}
    \label{fig:surgical_result}
\vspace{-2em}
\end{figure}

\subsection{Ablations}
\noindent
\textbf{Wavelet Function.} We analyze the impact of different wavelet functions on reconstruction quality, aiming to facilitate a comparison between our direction-aware representation and DWT wavelet. The evaluation is conducted on NSVF data \cite{liu2020neural}, where several complex wavelet functions with the approximate half-sample delay property—Antonini, LeGall, and two Near Symmetric filter banks (Near Symmetric A and Near Symmetric B)—are selected for comparison. Table \ref{tab:wavelet_function} reveals that the choice of different wavelets has minimal effect on reconstruction quality. Even the worst-performing wavelet function outperforms the discrete wavelet transform, underscoring the advantages of our direction-aware representation.
\begin{table}[t]
    \centering
    \caption{
    \textbf{Impact of Wavelet Transform Type/Function}, on reconstruction performance, evaluated on NSVF data..
    }
    \vspace{-.5em}
    \begin{tabular}{ccc}
    \toprule
        Wavelet Type&Wavelet Function & PSNR $\uparrow$ \\
        \midrule
        \multirow{4}*{DWT}& Haar&34.61\\
        ~& Coiflets 1 &34.56\\
        ~& \textbf{biorthogonal 4.4}&\textbf{34.67}\\
        ~& Daubechies 4 &34.44\\
        \midrule
         \multirow{4}*{DTCWT}&Antonini & 36.10\\
         ~&LeGall & 36.14\\
         ~&\textbf{Near Symmetric A}& \textbf{36.24}\\
         ~&Near Symmetric B& 36.17\\
         \bottomrule
    \end{tabular}
    \label{tab:wavelet_function}
\end{table}

\noindent
\textbf{Sparsity Analysis}. We evaluate the sparsity of our direction-aware representation by varying the sparsity level using different $\lambda_m$ values on the NSVF dataset. As depicted in Table \ref{tab:sparsity}, our direction-aware representation consistently achieves over 99\% sparsity. This remarkable sparsity, coupled with a model size of approximately 1MB, demonstrates the efficiency of our method in modeling static scenes 
while outperforming
state-of-the-art sparse representation methods.

\begin{table}[t]
    \centering
    \caption{
    \textbf{Sparsity Analysis of Direction-Aware Representation}, evaluated on NVSF data.}
    \vspace{-.5em}
    \resizebox{\linewidth}{!}{
    \begin{tabular}{cccc}
    \toprule
         $\lambda_m$&Sparsity $\uparrow$&Model Size (MB)  $\downarrow$ &PSNR $\uparrow$ \\
         \midrule
         $1.0\times 10^{-10}$&\textbf{99.2\%}& \textbf{1.16 MB}&35.36\\
         $5.0\times 10^{-11}$&97.3\%& 3.18 MB&35.81\\
         $2.5\times 10^{-11}$&94.2\%& 8.98 MB&\textbf{36.24}\\
         $0$&-&135 MB&36.34\\
         \bottomrule
    \end{tabular}
    }
    \label{tab:sparsity}
\vspace{-1.25em}
\end{table}

\noindent
\textbf{Wavelet Levels.} We investigated the impact of scene reconstruction performance across different wavelet levels, and the results are presented in supplementary material. We observed that increasing the wavelet level did not lead to significant performance improvements. Conversely, we noted a substantial increase in both training time and model size with the increment of wavelet level. As a result, throughout all experiments, we consistently set the wavelet level to 1.

\section{Conclusion}
We introduced a novel direction-aware representation capable of effectively capturing information from six different directions. The shift-invariant and direction-selective nature of our proposed representation enables the high-fidelity reconstruction of challenging dynamic scenes without requiring prior knowledge about the scene dynamics. Although this approach introduces some storage redundancy, we mitigate this by incorporating trainable masks for both static and dynamic scenes, resulting in a model size comparable to recent methods. Our proposed method is applicable to both NeRF and Gaussian Splatting settings for various types of dynamic scene reconstruction and demonstrates superior performance in recovering extremely fine details.

\section*{Acknowledgments}
This work was supported in part by NIH grant R01DC014037 and R01DC008408 from the National Institute of Deafness and Other Communications Disorders. The content is solely the responsibility of the authors and does not necessarily reflect the views of this institute. The authors would like to acknowledge Dr. Robert Labadie at MUSC for providing de-identified surgery videos used in this study.
\bibliography{ref}

\newpage

\section{Supplementary Material}

In this supplementary material, we provide further methodological context and implementation details to facilitate reproducibility of our framework DaReNeRF. We also showcase additional quantitative and qualitative results to further highlight the contributions claimed in the paper. 

\subsection{Video Presentation}
A video presentation of DaReNeRF and its results can be found online, at \url{https://www.youtube.com/watch?v=hYQsl6Rbxn4}.

\subsection{Dual-Tree Complex Wavelet Transform}
\begin{figure}[!h]
    \centering
    \includegraphics[width=1.\linewidth]{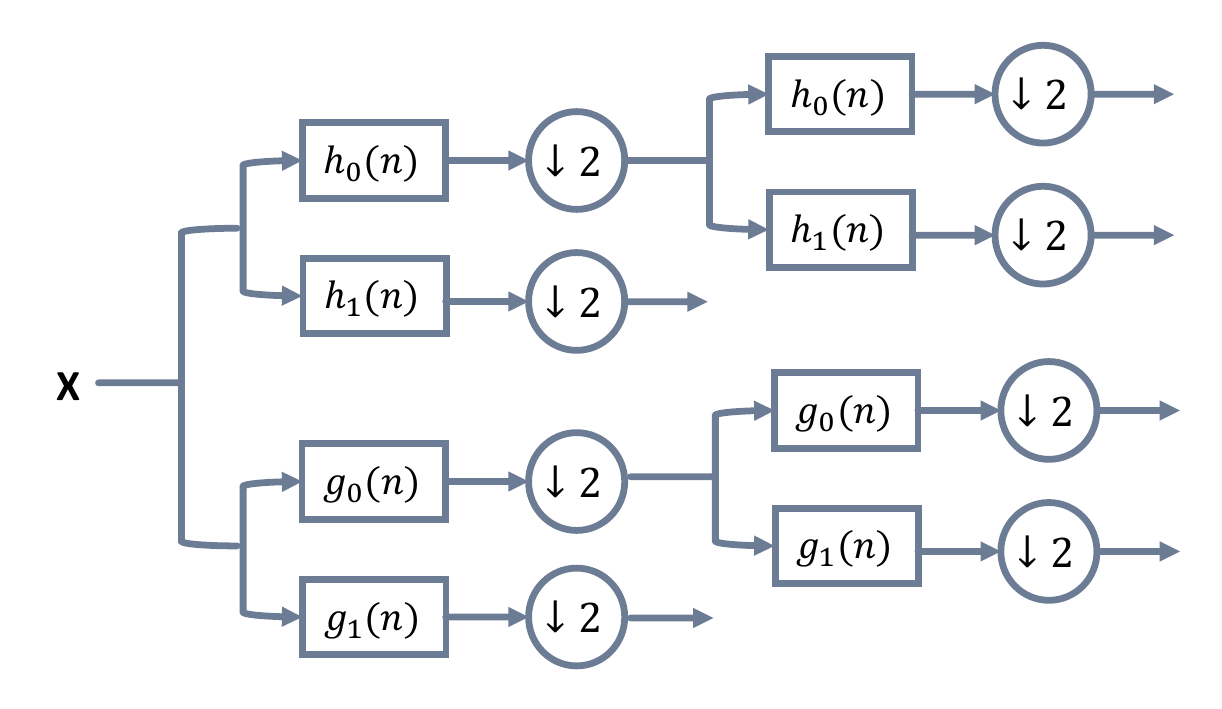}
    \caption{\textbf{Analysis filter bank}, for the dual tree complex wavelet transfrom.}
        \label{fig:DTCWT_filter_bank}
\end{figure}

The idea of dual-tree complex wavelet transform (DTCWT) \cite{selesnick2005dual} is quite straightforward. The DTCWT employs two real discrete wavelet transforms (DWTs). The first DWT gives the real part of the transform while the second DWT gives the imaginary part. The analysis filter banks used to implement the DTCWT is illustrated in Figure \ref{fig:DTCWT_filter_bank}. Here $h_0(n)$, $h_1(n)$ denote the low-pass/high-pass filter pair for upper filter bank, and $g_0(n)$, $g_1(n)$ denote the low-pass/high-pass filter pair for the lower filter bank. The two real wavelets associated with each of the two real wavelet transforms as $\psi_h(t)$ and $\psi_g(t)$. And the complex wavelet can be denoted as $\psi(t)=\psi_h(t)+j\psi_g(t)$. The $\psi_g(t)$ is approximately the Hilbert transform of $\psi_h(t)$. The 2D DTCWT $\psi(x,y) = \psi(x)\psi(y)$ associated with the row-column implementation of the wavelet transform, where $\psi(x)$ is a complex wavelet given by $\psi(x)=\psi_h(x)+j\psi_g(x)$. Then we obtain for $\psi(x,y)$ the expression:
\begin{small}
\begin{equation}
\begin{split}
    \psi(x,y)&=[\psi_h(x)+j\psi_g(x)][\psi_h(y)+j\psi_g(y)] \\
    &= \psi_h(x)\psi_h(y)-\psi_g(x)\psi_g(y) \\
    &+ j[\psi_g(x)\psi_h(y)+\psi_h(x)\psi_g(y)]
\end{split}
\label{euqation_1}
\end{equation}
\end{small}
The spectrum of $\psi_h(x)\psi_h(y)-\psi_g(x)\psi_g(y)$ which corresponds to the real part of $\psi(x,y)$ is supported in two quadrants of the 2D frequency plane and is oriented at $-45^\circ$. Note that the $\psi_h(x)\psi_h(y)$ is the HH wavelet of a separable 2D real wavelet transform implemented using the filter pair $\{h_0(n),h_1(n)\}$. Similarly, $\psi_g(x)\psi_g(y)$ is the HH wavelet of a real separable wavelet transform, implemented using the filters $\{g_0(n),g_1(n)\}$. To obtain a real 2D wavelet oriented at $+45^\circ$, we consider now the complex 2-D wavelet $\psi(x,y)=\psi(x)\overline{\psi(y)}$, where $\overline{\psi(y)}$ represents the complex conjugate of $\psi(y)$. This gives us the following expression:
\begin{small}
\begin{equation}
\begin{split}
    \psi(x,y)&=[\psi_h(x)+j\psi_g(x)][\overline{\psi_h(y)+j\psi_g(y)}] \\
    &= \psi_h(x)\psi_h(y)+\psi_g(x)\psi_g(y) \\
    &+ j[\psi_g(x)\psi_h(y)-\psi_h(x)\psi_g(y)]
\end{split}
\label{euqation_2}
\end{equation}
\end{small}
The spectrum of $\psi_h(x)\psi_h(y)+\psi_g(x)\psi_g(y)$ is supported in two quadrants of the 2D frequency plane and is oriented at $+45^\circ$. We could obtain four more oriented real 2D wavelets by repeating the above procedure on the following complex 2-D wavelets: $\phi(x)\psi(y)$, $\psi(x)\phi(y)$, $\phi(x)\overline{\psi(y)}$ and $\psi(x)\overline{\phi(y)}$, where $\psi(x) = \psi_h(x)+j\psi_g(y)$ and $\phi(x)=\phi_h(x)+j\phi_g(y)$. By taking the real part of each of these four complex wavelets, we obtain four real oriented 2D wavelets, in additional to the two already obtain in \ref{euqation_1} and \ref{euqation_2}:
\begin{small}
\begin{equation}
    \psi_i(x,y) = \dfrac{1}{\sqrt{2}}(\psi_{1,i}(x,y)-\psi_{2,i}(x,y)),
\label{euqation_3}
\end{equation}
\end{small}
\begin{small}
\begin{equation}
    \psi_{i+3}(x,y) = \dfrac{1}{\sqrt{2}}(\psi_{1,i}(x,y)+\psi_{2,i}(x,y))
\label{euqation_4}
\end{equation}
\end{small}
for $i=1,2,3$, where the two separable 2-D wavelet bases are defined in the usual manner:
\begin{small}
\begin{equation}
\begin{split}
    \psi_{1,1}(x,y)&=\phi_h(x)\psi_h(y), \psi_{2,1}(x,y)=\phi_g(x)\psi_g(y), \\
    \psi_{1,2}(x,y)&=\psi_h(x)\phi_h(y), \psi_{2,2}(x,y)=\psi_g(x)\phi_g(y), \\
    \psi_{1,3}(x,y)&=\psi_h(x)\psi_h(y), \psi_{2,3}(x,y)=\psi_g(x)\psi_g(y),
\end{split}
\end{equation}
\end{small}
We have used the normalization $\dfrac{1}{\sqrt{2}}$ only so that the sum and difference operation constitutes an orthonormal operation. From the imaginary parts of $\psi(x)\psi(y)$, $\psi(x)\overline{\psi(y)}$, $\phi(x)\psi(y)$, $\psi(x)\phi(y)$, $\phi(x)\overline{\psi(y)}$ and $\psi(x)\overline{\phi(y)}$, we could obtain six oriented wavelets given by:
\begin{small}
\begin{equation}
    \psi_i(x,y) = \dfrac{1}{\sqrt{2}}(\psi_{3,i}(x,y)+\psi_{4,i}(x,y)),
\end{equation}
\end{small}
\begin{small}
\begin{equation}
    \psi_{i+3}(x,y) = \dfrac{1}{\sqrt{2}}(\psi_{3,i}(x,y)-\psi_{4,i}(x,y))
\end{equation}
\end{small}
for $i=1,2,3$, where the two separable 2D wavelet bases are defined as:
\begin{small}
\begin{equation}
\begin{split}
    \psi_{3,1}(x,y)&=\phi_g(x)\psi_h(y), \psi_{4,1}(x,y)=\phi_h(x)\psi_g(y), \\
    \psi_{3,2}(x,y)&=\psi_g(x)\phi_h(y), \psi_{4,2}(x,y)=\psi_h(x)\phi_g(y), \\
    \psi_{3,3}(x,y)&=\psi_g(x)\psi_h(y), \psi_{4,3}(x,y)=\psi_h(x)\psi_g(y),
\end{split}
\end{equation}
\end{small}
Thus we could obtain six oriented wavelets from both real and imaginary part. 

\subsection{Additional Results on Various Datasets}

\begin{figure*}[t]
    \centering
  \animategraphics[width=\linewidth,controls,autoplay,loop,buttonsize=.7em,poster=last]{5}{img/plenoptic_anim/Slide}{1}{6}
  
    \caption{
    \textbf{Visual comparison on dynamic scenes (Plenoptic data).}
    K-Planes and HexPlane are concurrent decomposition-based methods. 
    As shown in the four zoomed-in patches, our method better reconstruct fine details and captures motion.
    To see the figure animated, please view the document with compatible software, \eg,  \textit{Adobe Acrobat} or \textit{KDE Okular.}
  }
    
    \label{fig:Visual_results_plenoptic_anim}
\vspace{-1em}
\end{figure*}

\subsubsection{Plenoptic Video Dataset \cite{li2022neural}}
The quantitative results for each scene are presented in Table \ref{tab:plenoptic_quantitative}, while additional visualizations comparing DaReNeRF with current state-of-the-art methods, HexPlane \cite{cao2023hexplane} and K-Planes \cite{fridovich2023k}, are provided in Figure \ref{fig:vis_comparisons}. Notably, DaReNeRF demonstrates superior recovery of texture details. 
We also provide an animated qualitative comparison in Figure \ref{fig:Visual_results_plenoptic_anim}.
Furthermore, comprehensive visualizations of DaReNeRF on all six scenes in the Plenoptic dataset are shown in Figure \ref{fig:fig_1_plenoptic} and Figure \ref{fig:fig_2_plenoptic}.

\begin{table*}[t]
    \centering
    \caption{Results on Plenoptic Video dataset. We report results of each scene}
    \resizebox{\linewidth}{!}{
    \begin{tabular}{cccccccccc}
    \toprule
         \multirow{2}*{Model}&\multicolumn{3}{c}{Flame Salmon}&\multicolumn{3}{c}{Cook Spinach}&\multicolumn{3}{c}{Cut Roasted Beef} \\
         ~&PSNR $\uparrow$ &D-SSIM $\downarrow$ &LPIPS $\downarrow$&PSNR $\uparrow$ &D-SSIM $\downarrow$ &LPIPS $\downarrow$&PSNR $\uparrow$ &D-SSIM $\downarrow$ &LPIPS $\downarrow$\\
    \midrule
    DaReNeRF-S&30.294&0.015&0.089&32.630&0.013&0.100&33.087&0.013&0.092\\
    \textbf{DaReNeRF}&\textbf{30.441}&\textbf{0.012}&\textbf{0.084}&\textbf{32.836}&\textbf{0.011}&\textbf{0.090}&\textbf{33.200}&\textbf{0.011}&\textbf{0.091}\\
    \midrule
    \multirow{2}*{ }&\multicolumn{3}{c}{Flame Steak}&\multicolumn{3}{c}{Sear Steak}&\multicolumn{3}{c}{Coffee Martini} \\
    DaReNeRF-S&33.259&0.011&0.081&33.179&0.011&0.075&30.160&0.016&0.092\\
    \textbf{DaReNeRF}&\textbf{33.524}&\textbf{0.009}&\textbf{0.077}&\textbf{33.351}&\textbf{0.009}&\textbf{0.072}&\textbf{30.193}&\textbf{0.014}&\textbf{0.089}\\
    \bottomrule
    \end{tabular}
    }
    \label{tab:plenoptic_quantitative}
\end{table*}

\begin{figure*}[t]
    \centering
    \includegraphics[width=\linewidth]{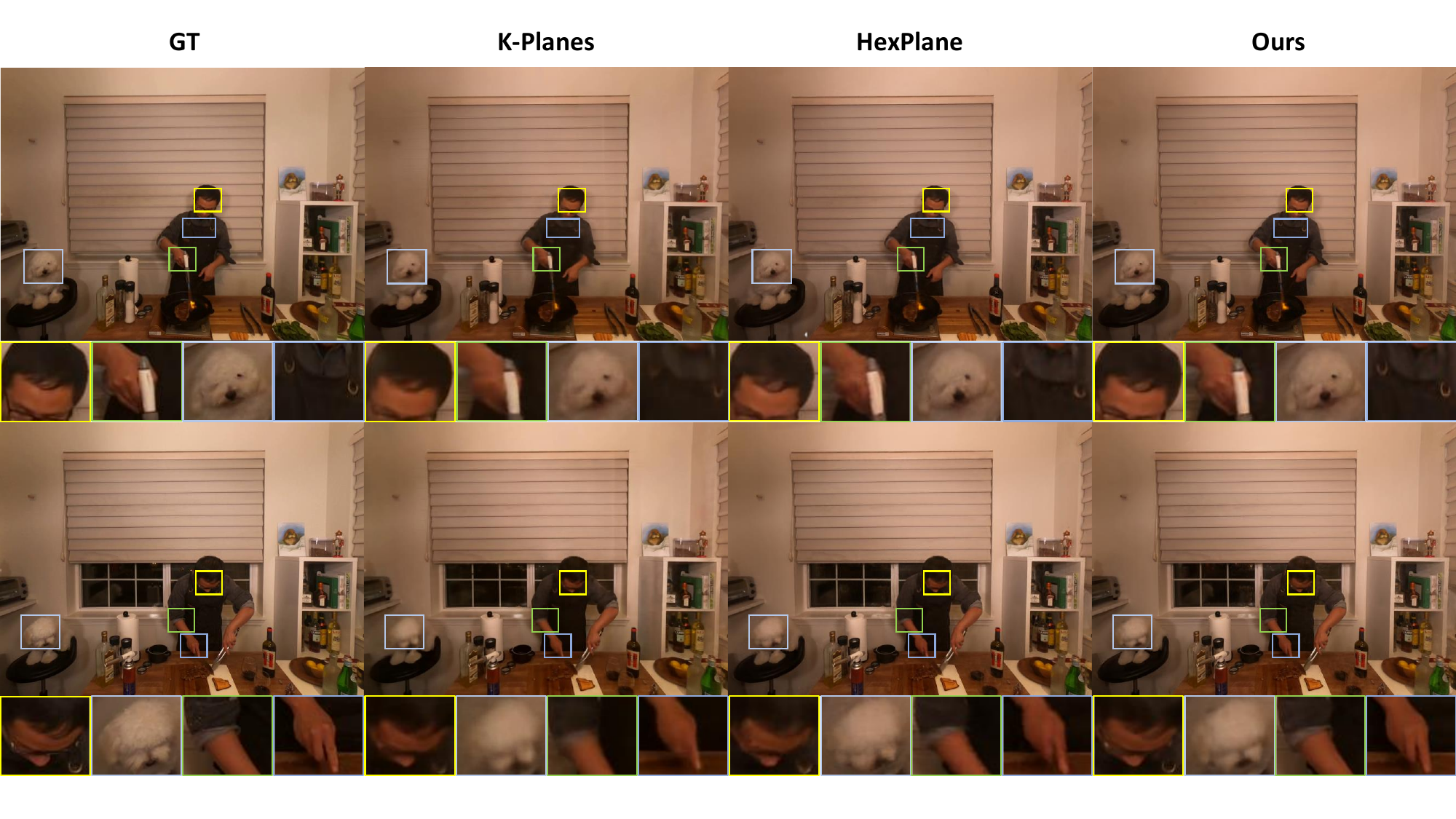}
    \caption{Visual comparison on dynamic scenes (Plenoptic data). K-Planes and HexPlane are concurrent decomposition-based methods.
As shown in the four zoomed-in patches, our method better reconstructs fine details and captures motion.}
    \label{fig:vis_comparisons}
\end{figure*}

\subsubsection{D-NeRF Dataset \cite{pumarola2021d}}
We provide quantitative results for each scene in Table \ref{tab:d_nerf_quan}, while additional visualizations comparing DaReNeRF with current state-of-the-art methods, HexPlane \cite{cao2023hexplane} and 4D-GS \cite{wu20234d}, are shared in Figure \ref{fig:dnerf_compare}. We also provide further visualization in a video attached to this supplementary material. 
Remarkably, although 4D-GS incorporates a deformation field, DaReNeRF still outperforms it in certain cases from the D-NeRF dataset. Furthermore, comprehensive visualizations of DaReNeRF on six scenes in the Plenoptic dataset are shown in Figure \ref{fig:dnerf_good} and the failure cases are shown in Figure \ref{fig:dnerf_fail}.

\begin{table*}[t]
    \centering
    \caption{Results of D-NeRF Dataset. We report results of each scene}
    \resizebox{\linewidth}{!}{
    \begin{tabular}{cccccccccc}
    \toprule
         \multirow{2}*{Model}&\multicolumn{3}{c}{Hell Warrior}&\multicolumn{3}{c}{Mutant}&\multicolumn{3}{c}{Hook} \\
         ~&PSNR $\uparrow$ &SSIM $\uparrow$ &LPIPS $\downarrow$&PSNR $\uparrow$ &SSIM $\uparrow$ &LPIPS $\downarrow$&PSNR $\uparrow$ &SSIM $\uparrow$ &LPIPS $\downarrow$\\
         \midrule
         T-NeRF&23.19&0.93&0.08&30.56&0.96&0.04&27.21&0.94&0.06  \\
         D-NeRF&25.02&0.95&0.06&31.29&0.97&0.02&29.25&0.96&0.11 \\
         TiNeuVox-S&27.00&0.95&0.09&31.09&0.96&0.05&29.30&0.95&0.07\\
         TiNeuVox-B&28.17&0.97&0.07&33.61&0.98&0.03&31.45&0.97&0.05\\
         HexPlane&24.24&0.94&0.07&33.79&0.98&0.03&28.71&0.96&0.05\\
         \midrule
         DaReNeRF-S&25.71&0.95&0.04&34.08&0.98&0.02&29.04&0.96&0.04\\
         DaReNeRF&25.82&0.95&0.04&34.17&0.98&0.01&28.96&0.96&0.04\\
         \midrule
         \multirow{2}*{ }&\multicolumn{3}{c}{Bouncing Balls}&\multicolumn{3}{c}{Lego}&\multicolumn{3}{c}{T-Rex} \\
         ~&PSNR $\uparrow$ &SSIM $\uparrow$ &LPIPS $\downarrow$&PSNR $\uparrow$ &SSIM $\uparrow$ &LPIPS $\downarrow$&PSNR $\uparrow$ &SSIM $\uparrow$ &LPIPS $\downarrow$\\
         \midrule
         T-NeRF&37.81&0.98&0.12&23.82&0.90&0.15&30.19&0.96&0.13\\
         D-NeRF&38.93&0.98&0.10&21.64&0.83&0.16&31.75&0.97&0.03\\
         TiNeuVox-S&39.05&0.99&0.06&24.35&0.88&0.13&29.95&0.96&0.06\\
         TiNeuVox-B&40.73&0.99&0.04&25.02&0.92&0.07&32.70&0.98&0.03\\
         HexPlane&39.69&0.99&0.03&25.22&0.94&0.04&30.67&0.98&0.03\\
         \midrule
         DaReNeRF-S&42.24&0.99&0.01&25.24&0.94&0.03&31.75&0.98&0.03\\
         DaReNeRF&42.26&0.99&0.01&25.44&0.95&0.03&32.21&0.98&0.02\\
         \midrule
         \multirow{2}*{ }&\multicolumn{3}{c}{Stand Up}&\multicolumn{3}{c}{Jumping Jacks}&\multicolumn{3}{c}{Average} \\
         ~&PSNR $\uparrow$ &SSIM $\uparrow$ &LPIPS $\downarrow$&PSNR $\uparrow$ &SSIM $\uparrow$ &LPIPS $\downarrow$&PSNR $\uparrow$ &SSIM $\uparrow$ &LPIPS $\downarrow$\\
         \midrule
         T-NeRF&31.24&0.97&0.02&32.01&0.97&0.03&29.51&0.95&0.08\\
         D-NeRF&32.79&0.98&0.02&32.80&0.98&0.03&30.50&0.95&0.07\\
         TiNeuVox-S&32.89&0.98&0.03&32.33&0.97&0.04&30.75&0.96&0.07\\
         TiNeuVox-B&35.43&0.99&0.02&34.23&0.98&0.03&32.64&0.97&0.04\\
         HexPlane&34.36&0.98&0.02&31.65&0.97&0.04&31.04&0.94&0.04\\
         \midrule
         DaReNeRF-S&34.47&0.98&0.02&31.99&0.97&0.03&31.82&0.97&0.03\\
         DaReNeRF&34.58&0.98&0.02&32.21&0.97&0.03&31.95&0.97&0.03\\
         \bottomrule

    \end{tabular}
    }
    \label{tab:d_nerf_quan}
\end{table*}

\subsubsection{NeRF Synthetic Dataset}
The quantitative results for each case are presented in Table \ref{tab:quant_nerf}, while additional visualizations comparing our representation with DWT \cite{rho2023masked} based representation method, are shown in Figure \ref{fig:nerf_synthetic_1}.  Furthermore, comprehensive visualizations of eight scenes in the NeRF dataset are shown in Figure \ref{fig:nerf_synthetic_2} and in the attached video.

\begin{table*}[t]
    \centering
    \caption{Results of NeRF Synthetic dataset.}
    \resizebox{\linewidth}{!}{
    \begin{tabular}{cccccccccccc}
    \toprule
         Bit Precision&Method&Size(MB)&Avg&Chair&Drums&Ficus&Hotdog&Lego&Materials&Mic&Ship  \\
         \midrule
         32-bit&KiloNeRF&$\leq$ 100&31.00&32.91&25.25&29.76&35.56&33.02&29.20&33.06&29.23\\
         32-bit&CCNeRF (CP)&4.4&30.55&-&-&-&-&-&-&-&-\\
         8-bit$^\ast$&NeRF&1.25&31.52&33.82&24.94&30.33&36.70&32.96&29.77&34.41&29.25\\
         8-bit&cNeRF&0.70&30.49&32.28&24.85&30.58&34.95&31.98&29.17&32.21&28.24\\
         8-bit$^\ast$&PREF&9.88&31.56&34.55&25.15&32.17&35.73&34.59&29.09&32.64&28.58\\
         8-bit$^\ast$&VM-192&17.93&32.91&35.64&25.98&33.57&37.26&36.04&29.87&34.33&30.64\\
         8-bit$^\ast$&VM-192 (300) + DWT&0.83&31.95&34.14&25.53&32.87&36.08&34.93&29.42&33.48&29.15\\
         \midrule
         8-bit$^\ast$&VM-192 (300) + Ours&8.91&32.42&36.05&29.40&35.26&36.37&25.58&33.26&29.82&33.63\\
         \bottomrule
         & 
    \end{tabular}
    }
    
    \label{tab:quant_nerf}
\end{table*}

\subsubsection{NSVF Synthetic Dataset}
The quantitative results for each case are presented in Table \ref{tab:quant_nsvf}, while additional visualizations comparing our representation with DWT \cite{rho2023masked} based representation method, are shown in Figure \ref{fig:nsvf_synthetic_1}.  Furthermore, comprehensive visualizations of eight scenes in the NSVF dataset are shown in Figure \ref{fig:nsvf_synthetic_2}.

\begin{table*}[t]
    \centering
    \caption{Results of NSVF synthetic dataset.}
    \resizebox{\linewidth}{!}{
    \begin{tabular}{cccccccccccc}
    \toprule
         Bit Precision&Method&Size(MB)&Avg&Bike&Lifestyle&Palace&Robot&Spaceship&Steamtrain&Toad&Wineholder  \\
         \midrule
         32-bit&KiloNeRF&$\leq$ 100&33.77&35.49&33.15&34.42&32.93&36.48&33.36&31.41&29.72\\
         8-bit$^\ast$&VM-192&17.77&36.11&38.69&34.15&37.09&37.99&37.66&37.45&34.66&31.16\\
         8-bit$^\ast$&VM-48&4.53&34.95&37.55&33.34&35.84&36.60&36.38&36.68&32.97&30.26\\
         8-bit$^\ast$&CP-384&0.72&33.92&36.29&32.29&35.73&35.63&34.58&35.82&31.24&29.75\\
         8-bit$^\ast$&VM-192 (300) + DWT&0.87&34.67&37.06&33.44&35.18&35.74&37.01&36.65&32.23&30.08\\
         \midrule
         8-bit$^\ast$&VM-192 (300) + Ours&8.98&36.24&38.78&34.21&37.22&38.02&38.61&37.79&34.39&30.97\\
         \bottomrule
         & 
    \end{tabular}
    }
    
    \label{tab:quant_nsvf}
\end{table*}

\subsubsection{LLFF Dataset}
The quantitative results for each case are presented in Table \ref{tab:quant_llff}, while additional visualizations comparing our representation with DWT \cite{rho2023masked} based representation method, are shown in Figure \ref{fig:llff_1}.  Furthermore, comprehensive visualizations of eight scenes in the NSVF dataset are shown in Figure \ref{fig:llff_2} and in the video.

\begin{table*}[t]
    \centering
    \caption{Results of LLFF dataset.}
    \resizebox{\linewidth}{!}{
    \begin{tabular}{cccccccccccc}
    \toprule
         Bit Precision&Method&Size(MB)&Avg&Fern&Flower&Fortress&Horns&Leaves&Orchids&Room&T-Rex  \\
         \midrule
         8-bit&cNeRF&0.96&26.15&25.17&27.21&31.15&27.28&20.95&20.09&30.65&26.72\\
         8-bit$^\ast$&PREF&9.34&24.50&23.32&26.37&29.71&25.24&20.21&19.02&28.45&23.67\\
         8-bit$^\ast$&VM-96&44.72&26.66&25.22&28.55&31.23&28.10&21.28&19.87&32.17&26.89\\
         8-bit$^\ast$&VM-48&22.40&26.46&25.27&28.19&31.06&27.59&21.33&20.03&31.70&26.54\\
         8-bit$^\ast$&CP-384&0.64&25.51&24.30&26.88&30.17&26.46&20.38&19.95&30.61&25.35\\
         8-bit$^\ast$&VM-96 (640) + DWT&1.34&25.88&24.98&27.19&30.28&26.96&21.21&19.93&30.03&26.45\\
         \midrule
         8-bit$^\ast$&VM-96 (640) + Ours&13.67&26.48&25.02&28.23&31.07&27.81&21.24&19.68&31.82&26.97\\
         \bottomrule
         & 
    \end{tabular}
    }
    
    \label{tab:quant_llff}
\end{table*}

\subsubsection{EndoNeRF Dataset \cite{wang2022neural}}
The quantitative results for two cases are present in Table \ref{tab:endonerf_results} and the visulization results are presented in the main paper.

\subsubsection{Hamlyn Dataset \cite{recasens2021endo}}
The quantitative results of 7 cases and average performance are shown in the Table \ref{tab:hamlyn_results}

\subsubsection{SCARED Dataset}
The quantitative results of 5 sequences and average performance are shown in the Table \ref{tab:scared_results}.

\subsubsection{Cochlear Implant Dataset}
The quantitative results of 2 cases are shown in the Table \ref{tab:cochlear_results}.

\subsection{Additional Ablation Studies}

\subsubsection{Sparsity Masks}
 We evaluate the performance of our direction-aware representation at various sparsity levels controlled by the mask loss weight $\lambda_m$. The quantitative and qualitative results on the NSVF dataset with different sparsity levels are presented in Table \ref{tab:abl_sparsity} and Figure \ref{fig:abl}.

 \begin{table*}[t]
    \centering
    \caption{Quantitative results on NSVF dataset with different sparsity.}
    \resizebox{\linewidth}{!}{
    \begin{tabular}{cccccccccccc}
    \toprule
         Sparsity&$\lambda_m$&Size(MB)&Avg&Bike&Lifestyle&Palace&Robot&Spaceship&Steamtrain&Toad&Wineholder  \\
         \midrule
         99.2\%&$1.0\times10^{-10}$&1.16&35.36&38.01&33.69&35.70&37.23&37.83&37.26&32.58&30.56\\
         97.3\%&$5.0\times10^{-11}$&3.18&35.81&38.52&34.01&36.33&37.79&38.22&37.46&33.33&30.82\\
         94.2\%&$2.5\times10^{-11}$&8.98&36.24&38.78&34.21&37.22&38.02&38.61&37.79&34.39&30.97\\
         -&0&135&36.34&38.86&34.37&37.25&38.06&38.72&37.89&34.46&31.09\\
         \bottomrule
         & 
    \end{tabular}
    }
    \label{tab:abl_sparsity}
\end{table*}

 \begin{table*}[t]
    \centering
    \caption{Quantitative results on EndoNeRF dataset.}
    \begin{tabular}{ccccccccc}
    \toprule
    \multirow{2}*{Model}&\multicolumn{4}{c}{Cutting}&\multicolumn{4}{c}{Pulling} \\
         ~&PSNR $\uparrow$ &SSIM $\uparrow$ &LPIPS $\downarrow$& Training Time (min) $\downarrow$&PSNR $\uparrow$ &SSIM $\uparrow$ &LPIPS $\downarrow$& Training Time (min) $\downarrow$\\
    \midrule
    ForPlane & 33.68& 0.900& 0.113& \textbf{4} &36.26&0.936&0.085&\textbf{4}\\
    DaReNeRF &\textbf{35.34}&\textbf{0.922}&\textbf{0.096}&5& \textbf{38.03}&\textbf{0.947}&\textbf{0.064}&5\\
    \midrule
    EndoGaussian & 38.10&0.962&0.047&\textbf{2.5}&37.00&0.957&0.070&\textbf{2.5}\\
    DaReGS &\textbf{38.38}&\textbf{0.965}&\textbf{0.031}&3.5&\textbf{38.32}&\textbf{0.967}&\textbf{0.049}&3.5\\
    \bottomrule
         
    \end{tabular}
    \label{tab:endonerf_results}
\end{table*}

 \begin{table*}[t]
    \centering
    \caption{Quantitative results on Hamlyn dataset.}
    \begin{tabular}{cccccccccc}
    \toprule
    \multirow{2}*{Model}&\multicolumn{3}{c}{Sequence 1}&\multicolumn{3}{c}{Sequence 2}&\multicolumn{3}{c}{Sequence 3} \\
         ~&PSNR $\uparrow$ &SSIM $\uparrow$ &LPIPS $\downarrow$& PSNR $\uparrow$ &SSIM $\uparrow$ &LPIPS $\downarrow$&PSNR $\uparrow$ &SSIM $\uparrow$ &LPIPS $\downarrow$\\
    \midrule
    ForPlane &32.86&0.919&0.124&33.88&0.932&0.125&33.66&0.933&0.123\\
    DaReNeRF &\textbf{33.12}&\textbf{0.926}&\textbf{0.121}&\textbf{34.39}&\textbf{0.941}&\textbf{0.113}&\textbf{34.01}&\textbf{0.938}&\textbf{0.119}\\
    \midrule
    \multirow{2}*{}&\multicolumn{3}{c}{Sequence 4}&\multicolumn{3}{c}{Sequence 5}&\multicolumn{3}{c}{Sequence 6} \\
         ~&PSNR $\uparrow$ &SSIM $\uparrow$ &LPIPS $\downarrow$& PSNR $\uparrow$ &SSIM $\uparrow$ &LPIPS $\downarrow$&PSNR $\uparrow$ &SSIM $\uparrow$ &LPIPS $\downarrow$\\
    \midrule
    ForPlane &37.89&0.963&0.075&38.90&0.971&0.041&35.24&0.945&0.077\\
    DaReNeRF &\textbf{38.65}&\textbf{0.971}&\textbf{0.058}&\textbf{39.29}&\textbf{0.973}&\textbf{0.038}&\textbf{35.90}&\textbf{0.959}&\textbf{0.065}\\
    \midrule
    \multirow{2}*{}&\multicolumn{3}{c}{Sequence 7}&\multicolumn{3}{c}{Averge}&\multicolumn{3}{c}{} \\
         ~&PSNR $\uparrow$ &SSIM $\uparrow$ &LPIPS $\downarrow$& PSNR $\uparrow$ &SSIM $\uparrow$ &LPIPS $\downarrow$&&&\\
    \midrule
    ForPlane &34.67&0.949&0.089&35.30&0.945&0.093&&&\\
    DaReNeRF &\textbf{35.12}&\textbf{0.956}&\textbf{0.085}&\textbf{35.64}&\textbf{0.952}&\textbf{0.085}&&&\\
   
    \bottomrule
         
    \end{tabular}
    \label{tab:hamlyn_results}
\end{table*}

 \begin{table*}[t]
    \centering
    \caption{Quantitative results on SCARED dataset.}
    \begin{tabular}{ccccccccc}
    \toprule
    \multirow{2}*{Model}&\multicolumn{4}{c}{Sequence 1}&\multicolumn{4}{c}{Sequence 2} \\
         ~&PSNR $\uparrow$ &SSIM $\uparrow$ &LPIPS $\downarrow$& Training Time (min) $\downarrow$&PSNR $\uparrow$ &SSIM $\uparrow$ &LPIPS $\downarrow$& Training Time (min) $\downarrow$\\
    \midrule
    EndoGaussian & 30.212& 0.870& 0.154& \textbf{2.5} &32.266&0.897&0.126&\textbf{2.5}\\
    DaReGS &\textbf{30.384}&\textbf{0.876}&\textbf{0.114}&3.5& \textbf{33.213}&\textbf{0.911}&\textbf{0.082}&3.5\\
    \midrule
    \multirow{2}*{}&\multicolumn{4}{c}{Sequence 3}&\multicolumn{4}{c}{Sequence 4} \\
         ~&PSNR $\uparrow$ &SSIM $\uparrow$ &LPIPS $\downarrow$& Training Time (min) $\downarrow$&PSNR $\uparrow$ &SSIM $\uparrow$ &LPIPS $\downarrow$& Training Time (min) $\downarrow$\\
    \midrule
    EndoGaussian & 19.523& 0.627& 0.533& \textbf{2.5} &25.819&0.868&0.373&\textbf{2.5}\\
    DaReGS &\textbf{20.551}&\textbf{0.646}&\textbf{0.468}&3.5& \textbf{26.336}&\textbf{0.870}&\textbf{0.337}&3.5\\
    \midrule
    \multirow{2}*{}&\multicolumn{4}{c}{Sequence 5}&\multicolumn{4}{c}{Average} \\
         ~&PSNR $\uparrow$ &SSIM $\uparrow$ &LPIPS $\downarrow$& Training Time (min) $\downarrow$&PSNR $\uparrow$ &SSIM $\uparrow$ &LPIPS $\downarrow$& Training Time (min) $\downarrow$\\
    \midrule
    EndoGaussian & 26.925& 0.874& 0.204& \textbf{2.5} &26.949&0.827&0.278&\textbf{2.5}\\
    DaReGS &\textbf{27.050}&\textbf{0.876}&\textbf{0.188}&3.5& \textbf{27.500}&\textbf{0.836}&\textbf{0.238}&3.5\\
    \bottomrule
         
    \end{tabular}
    \label{tab:scared_results}
\end{table*}

 \begin{table*}[t]
    \centering
    \caption{Quantitative results on Cochlear Implant dataset.}
    \begin{tabular}{ccccccccc}
    \toprule
    \multirow{2}*{Model}&\multicolumn{4}{c}{Case 1}&\multicolumn{4}{c}{Case 2} \\
         ~&PSNR $\uparrow$ &SSIM $\uparrow$ &LPIPS $\downarrow$& Training Time (min) $\downarrow$&PSNR $\uparrow$ &SSIM $\uparrow$ &LPIPS $\downarrow$& Training Time (min) $\downarrow$\\
    \midrule
    EndoGaussian & 33.985& 0.945& 0.057& \textbf{2.5} &34.063&0.950&0.072&\textbf{2.5}\\
    DaReGS &\textbf{34.422}&\textbf{0.953}&\textbf{0.072}&3.5& \textbf{34.350}&\textbf{0.951}&\textbf{0.065}&3.5\\  
    \bottomrule
    \end{tabular}
    \label{tab:cochlear_results}
\end{table*}

\begin{figure*}
    \centering
    \includegraphics[width=0.7\linewidth,height=\textheight]{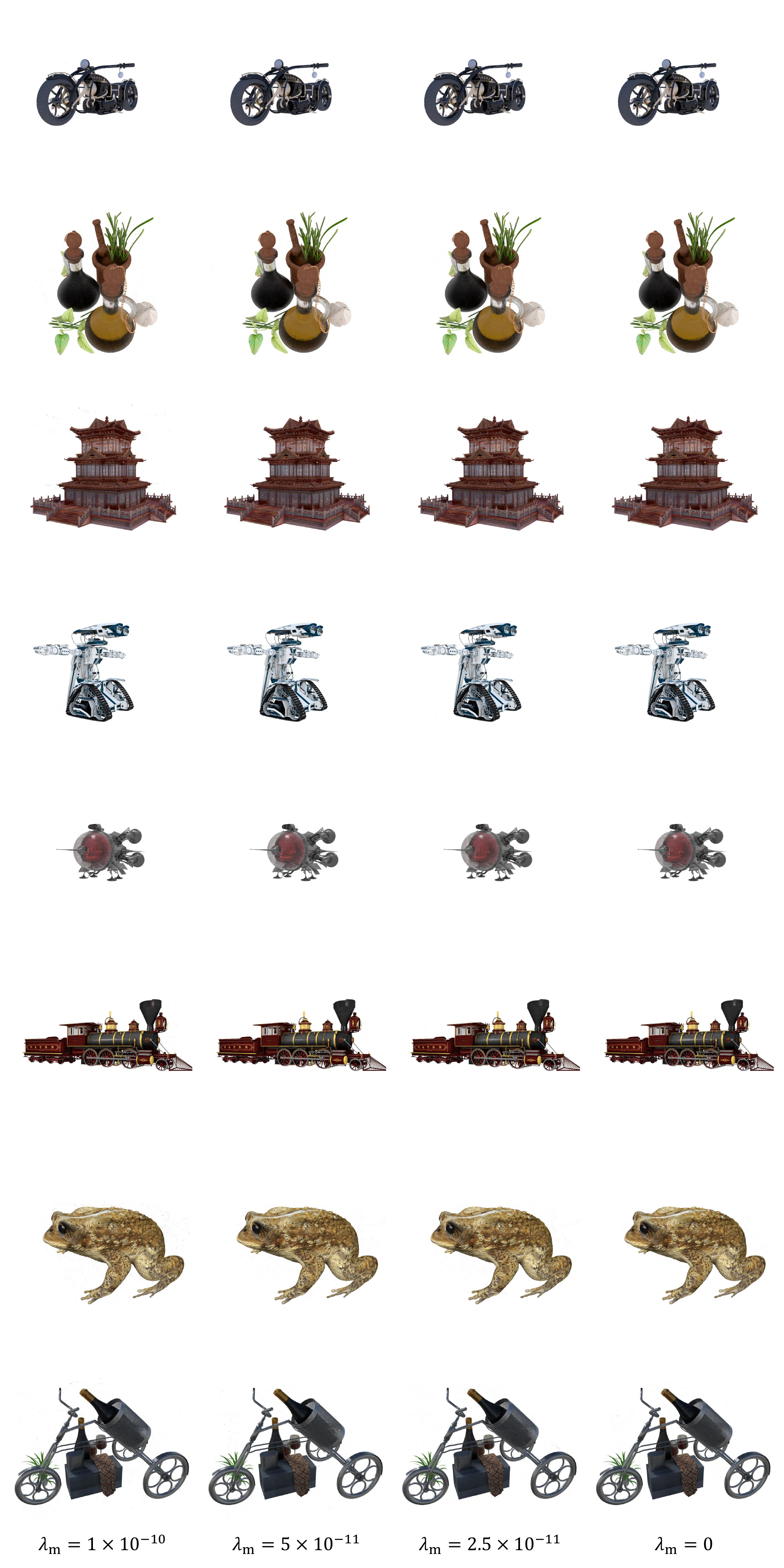}
    \caption{Qualitative results on NSVF dataset with different sparsity.}
    \label{fig:abl}
\end{figure*}

\subsubsection{Wavelet Levels} 
We investigated the impact of scene reconstruction performance across different wavelet levels, and the results are presented in Table \ref{tab:wavelet_level}. Interestingly, we observed that increasing the wavelet level did not lead to significant performance improvements. Conversely, we noted a substantial increase in both training time and model size with the increment of wavelet level. As a result, throughout all experiments, we consistently set the wavelet level to 1.

\begin{table}[t]
    \centering
    \caption{
    \textbf{Wavelet-level analysis of direction-aware representation}, evaluated on NVSF data.}
    \resizebox{\linewidth}{!}{
    \begin{tabular}{cccc}
    \toprule
         Level&PSNR $\uparrow$&Model Size (MB)  $\downarrow$ &Training Time (min) $\downarrow$ \\
         \midrule
         1&36.34& \textbf{135}&\textbf{23}\\
         2&36.45&152&41\\
         3&\textbf{36.49}& 163&55\\
         \bottomrule
    \end{tabular}
    }
    \label{tab:wavelet_level}
\end{table}

\subsubsection{Training Data Sparsity Analysis} 
In order to delve deeper into the few-shot capabilities of our proposed direction-aware representation, we conducted experiments with varying levels of training data sparsity. This was achieved by randomly dropping training frames while ensuring sufficient data remained to effectively learn motion on the D-NeRF dataset. The corresponding results are presented in Table \ref{tab:sparsity_of_training_set}. Remarkably, our proposed DaReNeRF consistently outperforms the baseline across different levels of training data sparsity.
\begin{table}[t]
    \centering
    \setlength{\tabcolsep}{2.0mm}
    \caption{Evaluation on D-NeRF with various training set sparsity.}
    \resizebox{\linewidth}{!}{
    \begin{tabular}{c|ccc|ccc}
    \toprule
         \multirow{2}*{Model}&\multicolumn{3}{c|}{\textbf{75\%} training set (average)} &\multicolumn{3}{c}{\textbf{50\%} training set (average)} \\
         ~&PSNR $\uparrow$ &SSIM $\uparrow$ &LPIPS $\downarrow$ &PSNR $\uparrow$ &SSIM $\uparrow$ &LPIPS $\downarrow$\\
        \midrule
        HexPlane&29.85&0.95&0.05&28.03&0.94&0.06\\
        DaReNeRF&\textbf{30.95}&\textbf{0.96}&\textbf{0.04}&\textbf{29.28}&\textbf{0.96}&\textbf{0.05}\\

    \bottomrule
    \end{tabular}
    }
    \label{tab:sparsity_of_training_set}
\end{table}

\subsection{Training Details}
\subsubsection{Plenoptic Video Dataset \cite{li2022neural}}
Plenoptic Video Dataset is a multi-view real-world video dataset, where each video is 10-second long. For training, we set $R_1=48$, $R_2=48$ and $R_3=48$ for appearance, where $R_1$, $R_2$ and $R_3$ are basis numbers for direction-aware representation of $XY-ZT$, $XZ-YT$ and $YZ-XT$ planes. For opacity, we set $R_1=24$, $R_2=24$ and $R_3=24$. The scene is modeled using normalized device coordinate (NDC) \cite{mildenhall2021nerf} with min boundaries $[-2.5,-2.0,-1.0]$ and max boundaries $[2.5,2.0,1.0]$. 

During the training, DaReNeRF starts with a space grid size of $64^3$ and double its resolution at 20k, 40k and 70k to $512^3$. The emptiness voxel is calculated at 30k, 50k and 80k. The learning rate for representation planes is 0.02 and the learning rate for $V^{RF}$ and neural network is 0.001. All learning rates are exponentially decayed. We use Adam \cite{kingma2014adam} optimization with $\beta_1=0.9$ and $\beta_2=0.99$. We apply the total variational loss on all representation planes with loss weight $\lambda=1e-5$ for spatial planes and $\lambda=2e-5$ for spatial-temporal planes. For DaReNeRF-S we set weight of mask loss as $1e-11$. 

We follow the hierarchical training pipeline suggested in \cite{li2022neural}. Both DaReNeRF and DaReNeRF-S use 100k iterations, with 10k stage one training, 50k stage two training and 40k stage three training. Stage one is a global-median-based weighted sampling with $\gamma=0.02$; stage two is also a global-median based weighted sampling with $\gamma=0.02$; stage three is a temporal-difference-based weighted sampling with $\gamma=0.2$.

In evaluation, D-SSIM is computed as $\dfrac{1-MS-SSIM}{2}$ and LPIPS \cite{zhang2018unreasonable} is calculated using AlexNet \cite{krizhevsky2012imagenet}. 

\subsubsection{D-NeRF Dataset \cite{pumarola2021d}}
We set $R_1=48$, $R_2=48$ and $R_3=48$ for appearance and $R_1=24$, $R_2=24$ and $R_3=24$ for opacity. The bounding box has max boundaries $[1.5,1.5,1.5]$ and min boundaries $[-1.5,-1.5,-1.5]$. During the training, both DaReNeRF and DaReNeRF-S starts with space grid of $32^3$ and upsampling its resolution at 3k, 6k and 9k to $200^3$. The emptiness voxel is calculated at 4k, 8k and 10k iterations. Total training iteration is 25k. The learning rate for representation planes are 0.02 and learning rate for $V^{RF}$ and neural network is 0.001. All learning rates are exponentially decayed. We use Adam \cite{kingma2014adam} optimization with $\beta_1=0.9$ and $\beta_2=0.99$. In evaluation, LPIPS \cite{zhang2018unreasonable} is calculated using VGG-Net  \cite{simonyan2014very} following previous works. 

For \textbf{both} the Plenoptic Video dataset and the D-NeRF dataset, we set the learning rate of the masks in DaReNeRF-S same as their representation planes and we employ a compact MLP for regressing output colors. The MLP consists of 3 layers, with a hidden dimension of 128.

\subsubsection{Static Scene}
For three static scene datasets NeRF synthetic dataset, NSVF synthetic dataset and LLFF dataset, we followed the experimental settings of TensoRF \cite{chen2022tensorf}. We trained our model for 30000 iterations, each of which is a minibatch of 4096 rays. We used Adam \cite{kingma2014adam} optimization with $\beta_1=0.9$ and $\beta_2=0.99$ and an exponential learning rate decay scheduler. The initial learning rates of representation-related parameters and neural network (MLP) were set to 0.02 and 0.001. For the \textbf{NeRF synthetic} and \textbf{NSVF synthetic} datasets, we adopt TensoRF-192 as the baseline and update the alpha masks at the 2k, 4k, 6k, 11k, 16k, and 26k iterations. The initial grid size is set to $128^3$, and we perform upsampling at 2k, 3k, 4k, 5.5k, and 7k iterations, reaching a final resolution of $300^3$. For the \textbf{LLFF} dataset, we adopt TensoRF-96 as the baseline and update the alpha masks at the 2.5k, 4k, 6k, 11k, 16k, and 21k iterations. The initial grid size is set to $128^3$, and we perform upsampling at 2k, 3k, 4k and 5.5k iterations, reaching a final resolution of $640^3$. The learning rates of masks are set same as learning rates of representation-related parameters. We employ a compact MLP for regressing output colors. The MLP consists of 3 layers, with a hidden dimension of 128.

\subsubsection{Dynamic Surgical Scene}
For the surgical dynamic scene datasets, we test our DaReNeRF and DaReGS models, following the settings of ForPlane \cite{yang2024efficient} and EndoGaussian \cite{liu2024endogaussian}, respectively.

\noindent
\textbf{DaReNeRF.} The surgical scene is normalized into normalized device coordinates (NDC), and the video duration is normalized to $[-1,1]$. The dimensions for the two stages of point sampling by a $Sample-Net$ \cite{yang2024efficient} are 128 and 256, respectively. A one-blob \cite{muller2019neural} encoding is applied to encode the spatiotemporal information. The full model employs a multi-resolution strategy with spatial axis resolutions of ${64, 128, 256, 512}$, while the temporal size is fixed at 100. The basis number for all spatiotemporal planes is set to 32. An Adam optimizer with an initial learning rate of 0.01 and a batch size of 2048 is selected for training.

\noindent
\textbf{DaReGS.} We randomly sample 0.1\% of the points as initialization points and select Adam as the optimizer with an initial learning rate of $1.6 \times 10^{-3}$. A warmup strategy is employed, where the Gaussian is first optimized for 1k iterations, followed by optimization of the entire framework for 3k iterations.

\subsection{Future works}
Future work could explore incorporating various frequency-based representations \cite{li2024neurobolt,li2024leveraging} and integrating segmentation masks \cite{lou2022caranet,lou2021dc,lou2023caranet,lou2023min,lou2024zero} with high-quality depth maps \cite{lou2024ws,lou2024surgical}. This approach holds potential to significantly accelerate training and improve the accuracy of reconstruction.

\clearpage

\begin{figure*}[t]
    \centering
    \includegraphics[width=\linewidth,height=\textheight]{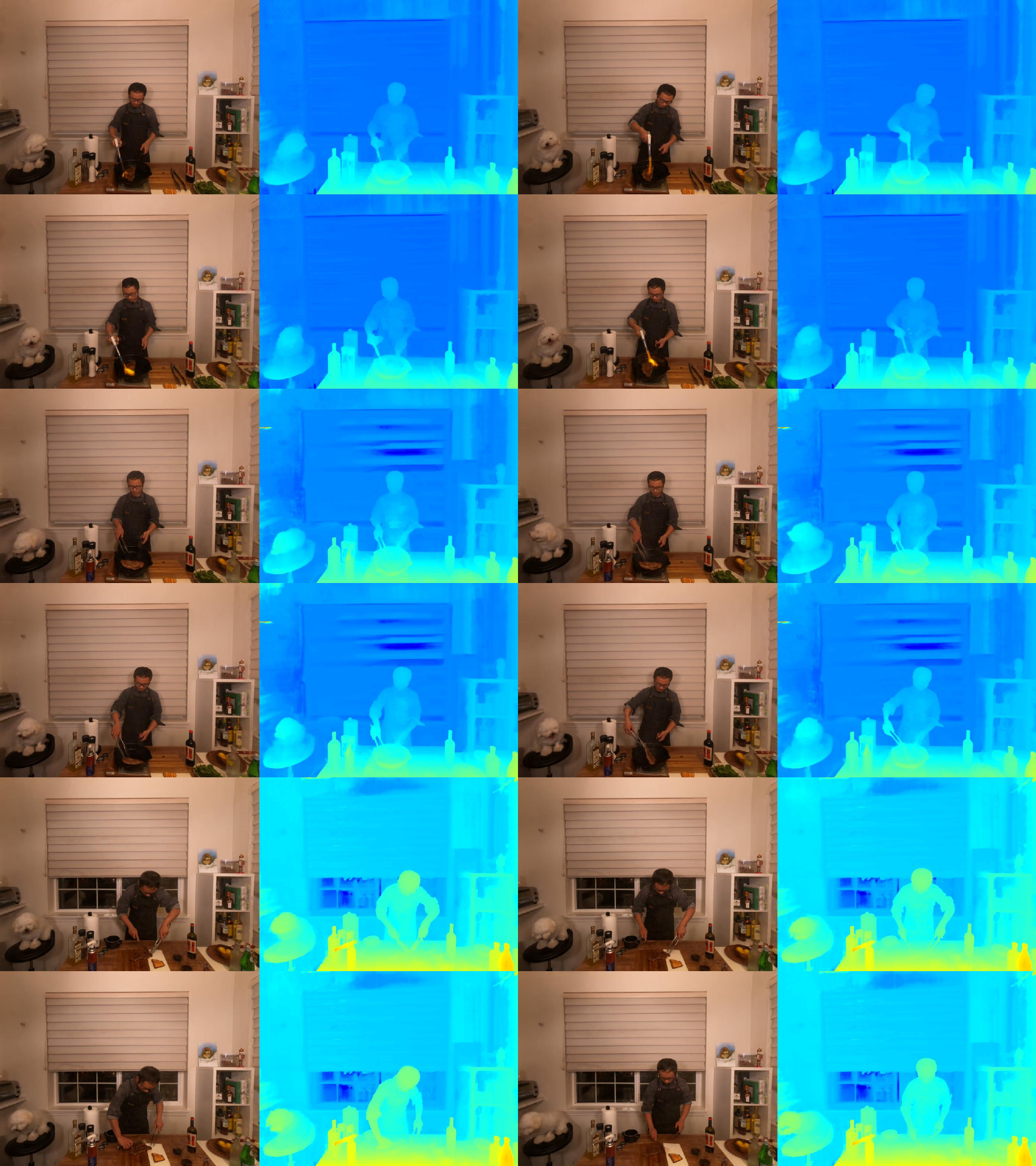}
    \caption{Visualizations on \texttt{flame steak}, \texttt{sear steak} and \texttt{cut roasted beef} scenes.}
    \label{fig:fig_1_plenoptic}
\end{figure*}

\begin{figure*}[t]
    \centering
    \includegraphics[width=\linewidth,height=\textheight]{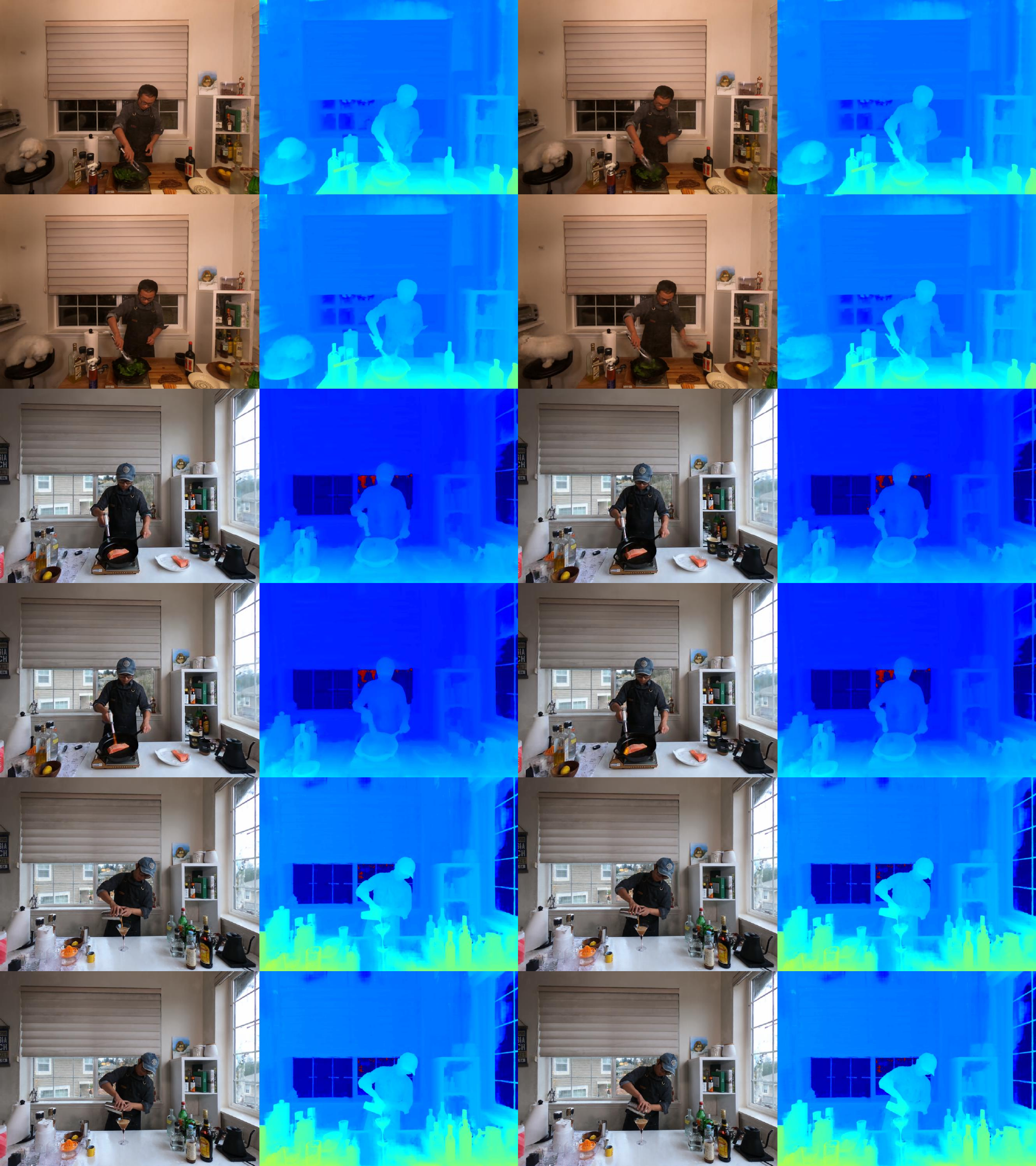}
    \caption{Visualizations on \texttt{cook spinach}, \texttt{flame salmon} and \texttt{coffee martini} scenes.}
    \label{fig:fig_2_plenoptic}
\end{figure*}

\begin{figure*}[t]
    \centering
    \includegraphics[width=\linewidth]{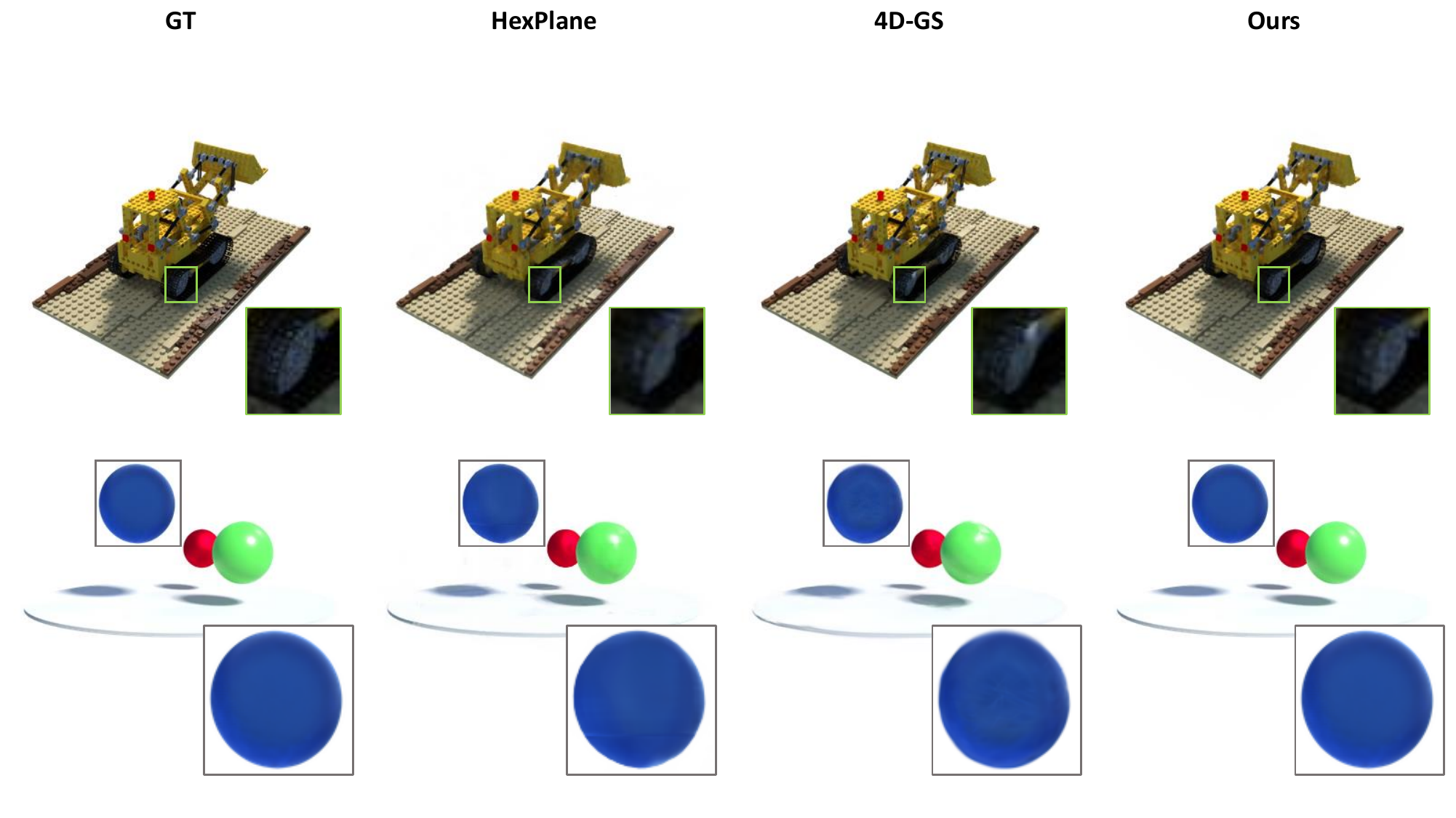}
    \caption{Visual Comparison on Dynamic Scenes (D-NeRF Data). 4D-GS and HexPlane are decomposition-based and deformation-based methods.}
    \label{fig:dnerf_compare}
\end{figure*}

\begin{figure*}[t]
    \centering
    \includegraphics[width=\linewidth,height=\textheight]{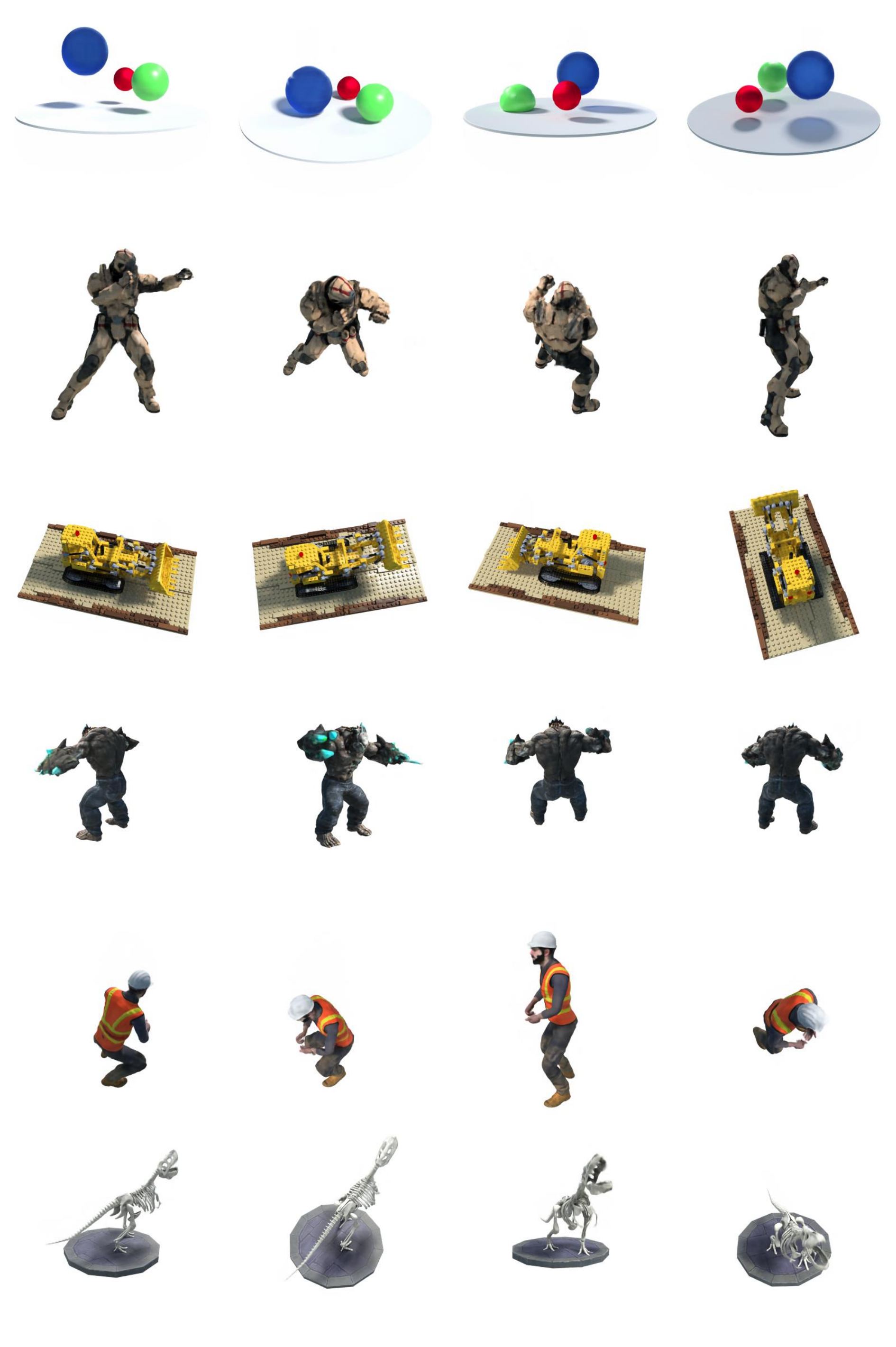}
    \caption{Visualizations on D-NeRF dataset.}
    \label{fig:dnerf_good}
\end{figure*}

\begin{figure*}[t]
    \centering
    \includegraphics[width=\linewidth]{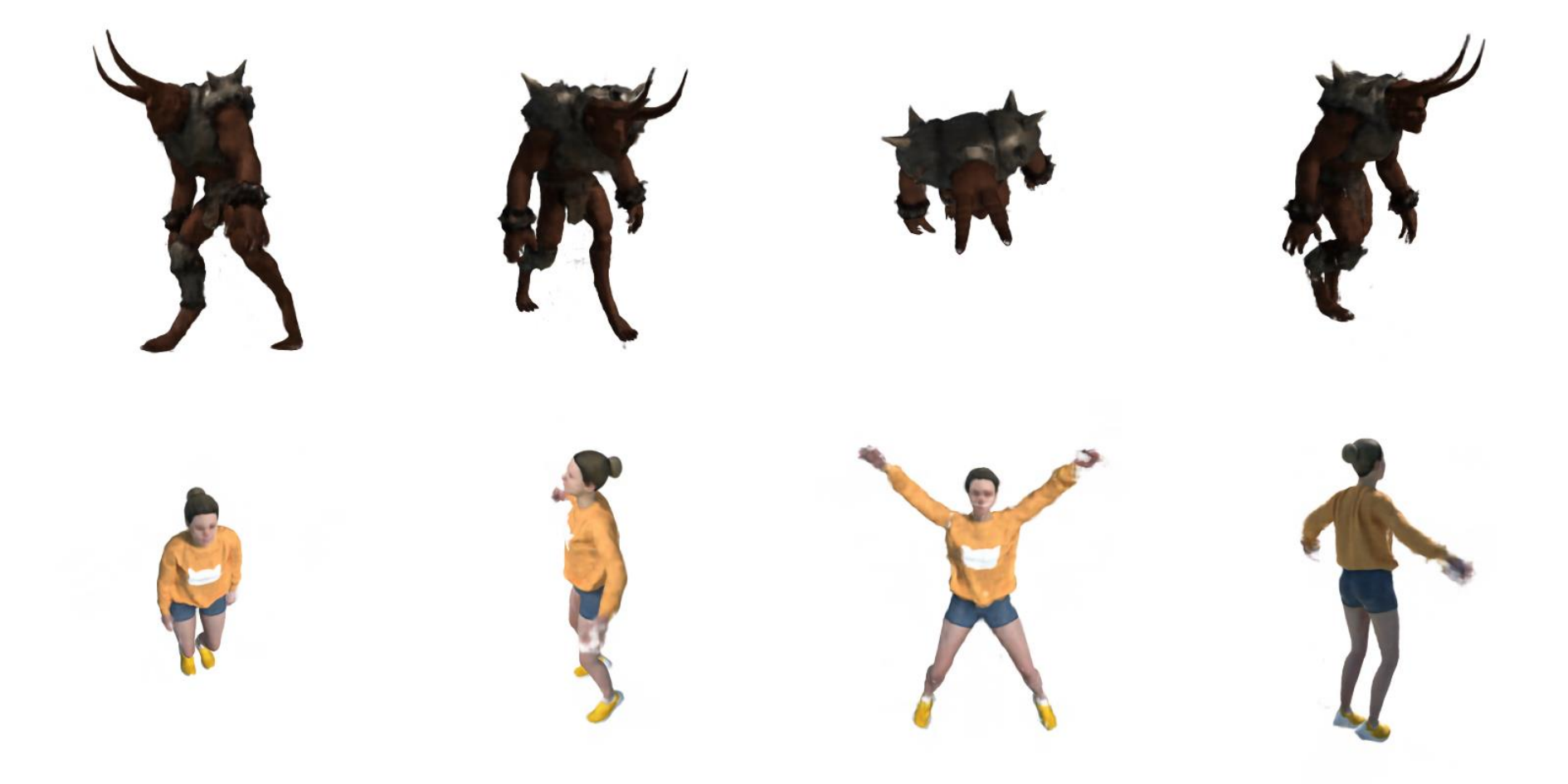}
    \caption{Visualizations on failure cases from D-NeRF dataset.}
    \label{fig:dnerf_fail}
\end{figure*}

\begin{figure*}
    \centering
    \includegraphics[width=\linewidth]{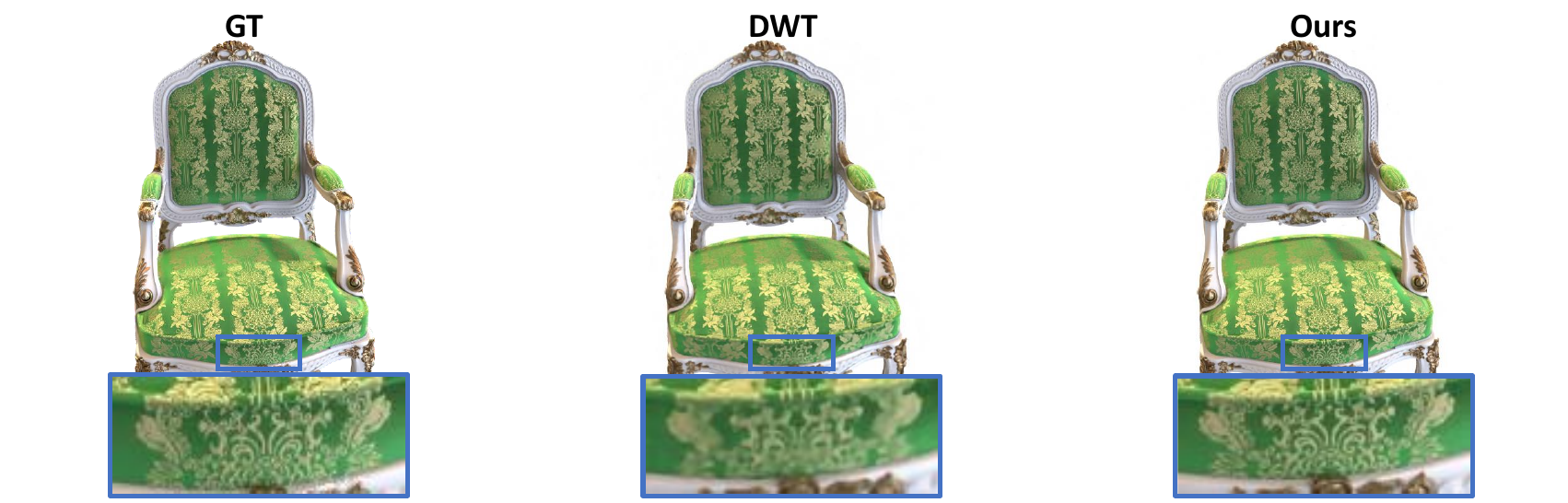}
    \caption{Visual comparison on NeRF synthetic dataset.}
    \label{fig:nerf_synthetic_1}
\end{figure*}

\begin{figure*}
    \centering
    \includegraphics[width=0.7\linewidth,height=\textheight]{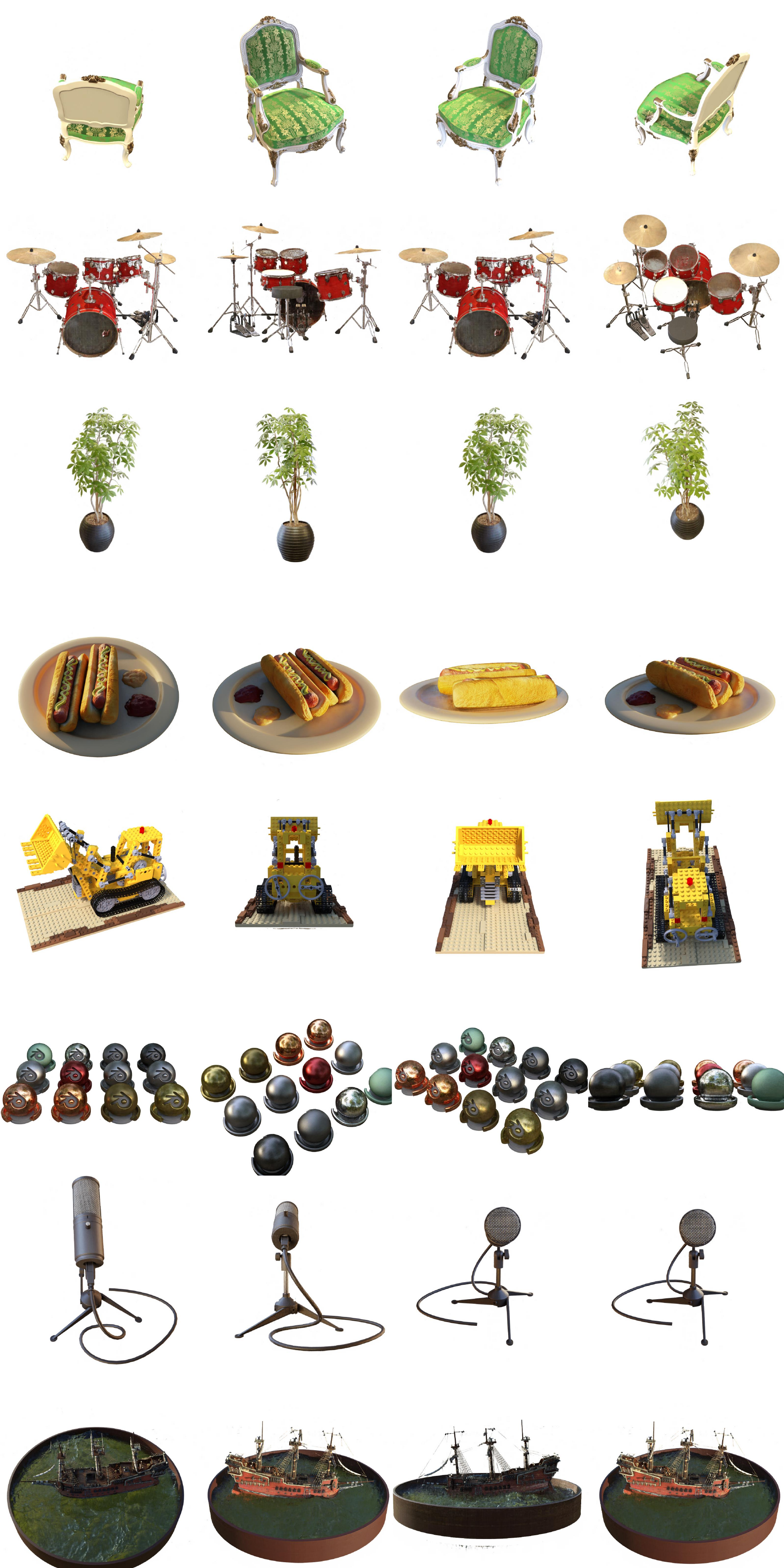}
    \caption{Visualizations on NeRF synthetic dataset.}
    \label{fig:nerf_synthetic_2}
\end{figure*}

\begin{figure*}
    \centering
    \includegraphics[width=\linewidth]{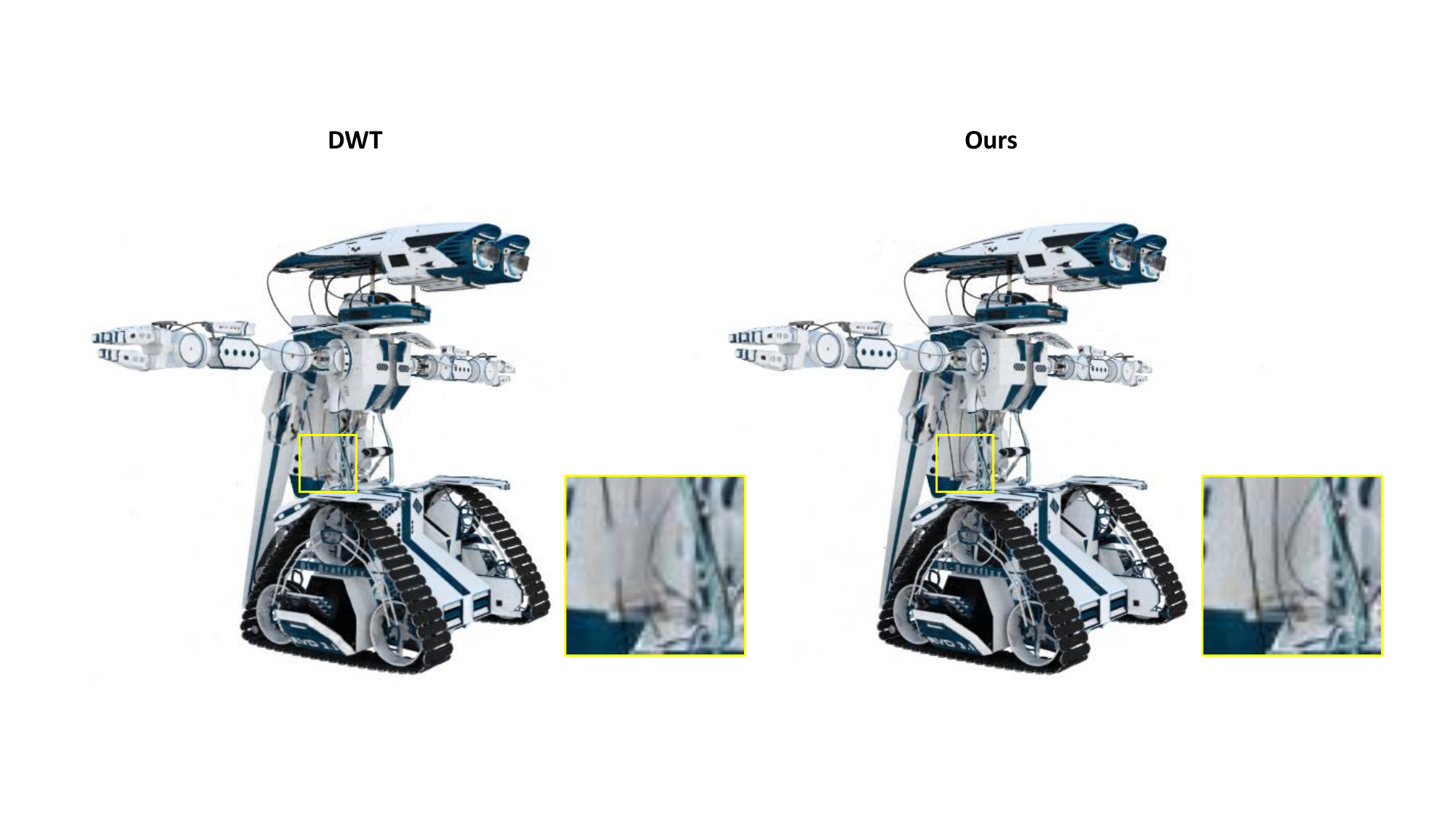}
    \caption{Visual comparison on NSVF synthetic dataset.}
    \label{fig:nsvf_synthetic_1}
\end{figure*}

\begin{figure*}
    \centering
    \includegraphics[width=0.7\linewidth,height=\textheight]{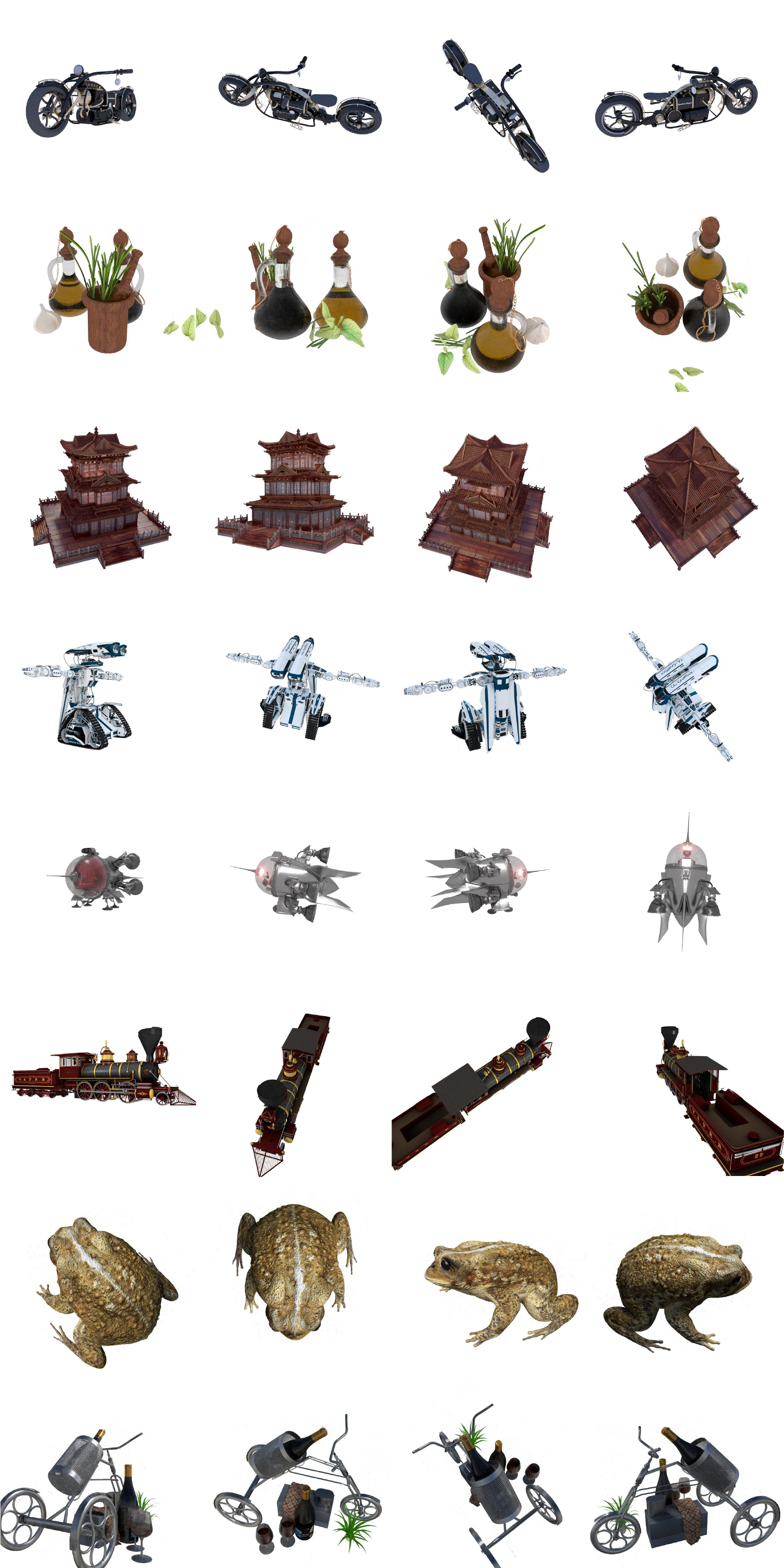}
    \caption{Visualizations on NSVF synthetic dataset.}
    \label{fig:nsvf_synthetic_2}
\end{figure*}

\begin{figure*}
    \centering
    \includegraphics[width=\linewidth]{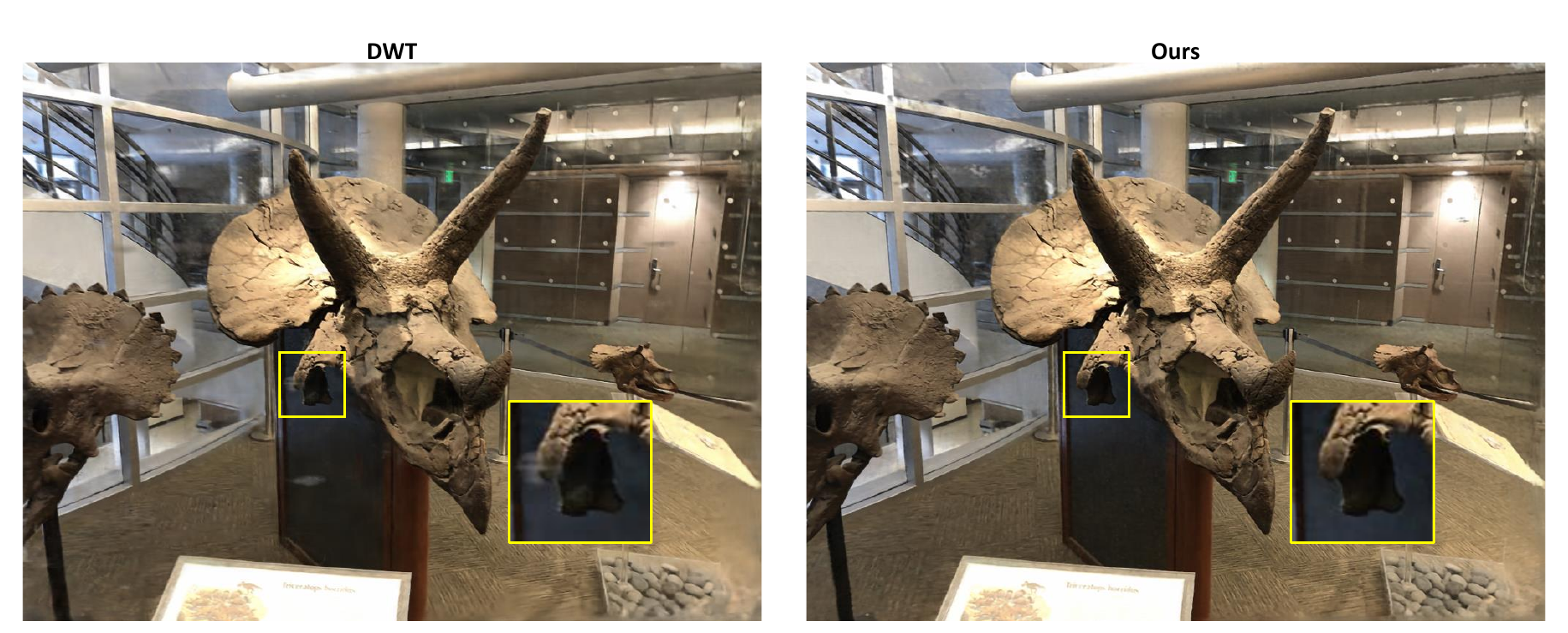}
    \caption{Visual comparison on LLFF synthetic dataset.}
    \label{fig:llff_1}
\end{figure*}

\begin{figure*}
    \centering
    \includegraphics[width=0.7\linewidth,height=\textheight]{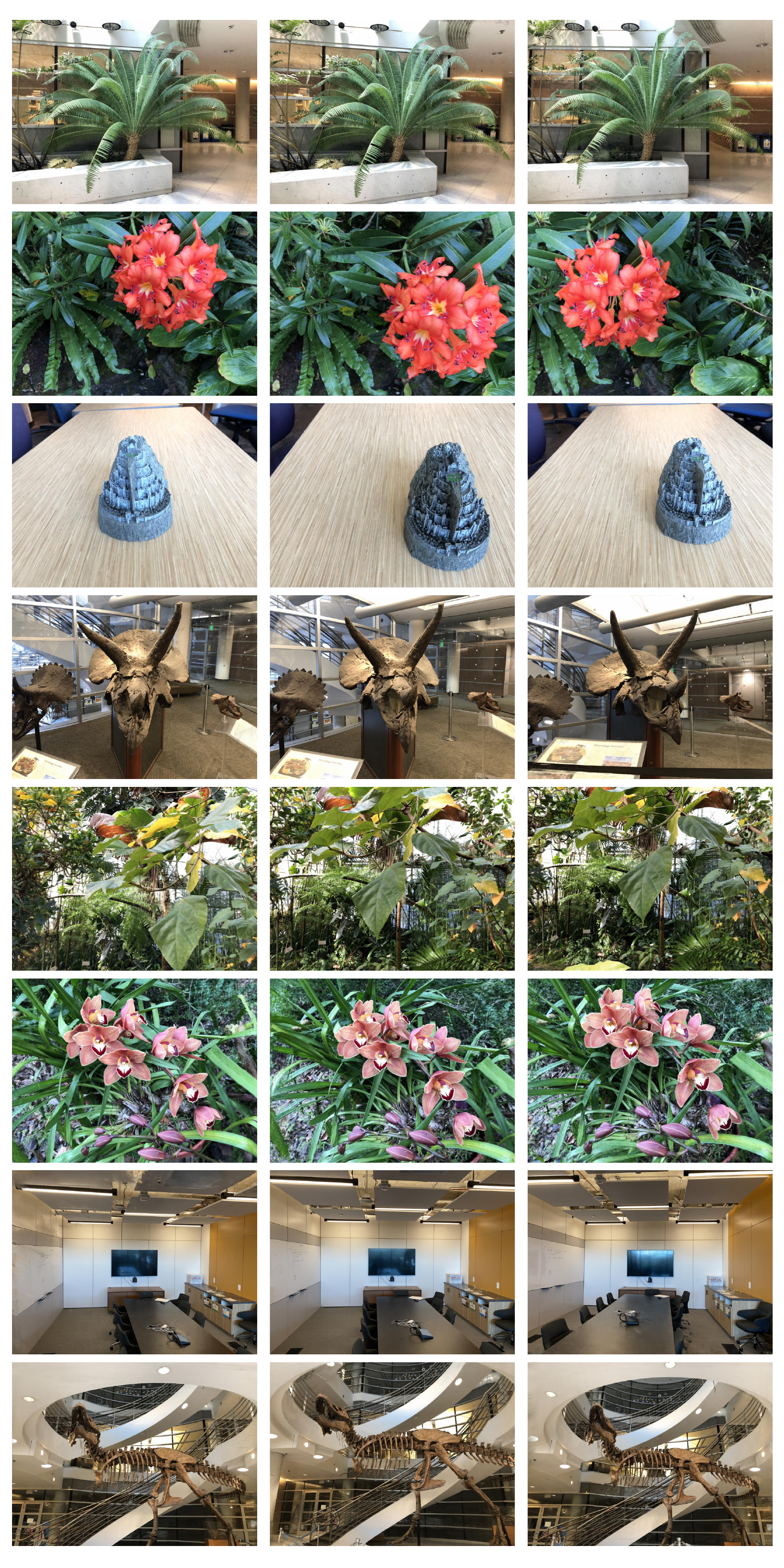}
    \caption{Visualizations on LLFF synthetic dataset.}
    \label{fig:llff_2}
\end{figure*}

\end{document}